\documentclass{article}

\usepackage[final,nonatbib]{nips_2016}

\usepackage[utf8]{inputenc} % allow utf-8 input
\usepackage[T1]{fontenc}    % use 8-bit T1 fonts
\usepackage{url}            % simple URL typesetting
\usepackage{booktabs}       % professional-quality tables
\usepackage{amsfonts}       % blackboard math symbols
\usepackage{nicefrac}       % compact symbols for 1/2, etc.
\usepackage{microtype}      % microtypography
\usepackage{amsmath}
\usepackage{graphicx}
\usepackage{multirow}
\usepackage{array}

\def\x{{\mathbf x}}
\def\y{{\mathbf y}}
\def\w{{\mathbf w}}
\def\z{{\mathbf z}}

\def\W{{\mathbf W}}
\def\V{{\mathbf V}}
\def\A{{\mathbf A}}
\def\B{{\mathbf B}}
\def\C{{\mathbf C}}
\def\Q{{\mathbf Q}}

\def\S{{\mathbf S}}
\def\U{{\mathbf U}}
\def\u{{\mathbf u}}
\def\Z{{\mathbf Z}}
\def\P{{\mathbf P}}
\def\E{{\mathbf E}}
\def\I{{\mathbf I}}
\def\Real{{\mathbb R}}

\newcommand\vsp{\vspace*{-0.1cm}}
\newcommand\Xcal{\mathcal X}

\newcommand\Sbb{\mathbb S}
\newcommand\Hcal{\mathcal H}
\newcommand\Zcal{\mathcal Z}

\newcommand\Ecal{\mathcal E}
\newcommand\Fcal{\mathcal F}
\newcommand\eg{\emph{e.g.}}
\newcommand\jmone{{j-1}}
\DeclareMathOperator*{\argmin}{arg\,min}
\newcommand{\defin}{:=}
\def\alphab{\boldsymbol\alpha}
\def\varepsilonb{\boldsymbol\varepsilon}
\newtheorem{lemma}{Lemma}
\newtheorem{proposition}{Proposition}
\let\originalleft\left
\let\originalright\right
\renewcommand{\left}{\mathopen{}\mathclose\bgroup\originalleft}
\renewcommand{\right}{\aftergroup\egroup\originalright}
\newcommand\proj{\text{Proj}_{\|.\|_2 = 1}}
\newcommand\projb{\text{Proj}_{\tilde{\Sbb}^{p-1}}}

\title{End-to-End Kernel Learning with \\ Supervised Convolutional Kernel Networks}

\author{
  Julien Mairal\\
  Inria\thanks{Thoth team, Inria Grenoble, Laboratoire Jean Kuntzmann, CNRS, Univ. Grenoble Alpes, France.}\\
  \texttt{julien.mairal@inria.fr} 
}

\begin{document}

\maketitle

\begin{abstract}
   In this paper, we introduce a new image representation based on a multilayer
kernel machine. Unlike traditional kernel methods where data representation
is decoupled from the prediction task, we learn how to shape the kernel with
supervision. We proceed by first proposing improvements of the
recently-introduced convolutional kernel networks (CKNs) in the context of
unsupervised learning; then, we derive backpropagation rules to take advantage
of labeled training data. The resulting model is a new type of convolutional
neural network, where optimizing the filters at each layer is equivalent to learning a
linear subspace in a reproducing kernel Hilbert space (RKHS). 
We show that our method achieves reasonably competitive performance for image classification on some standard
``deep learning'' datasets such as CIFAR-10 and
SVHN, and also for image super-resolution, demonstrating the applicability of
our approach to a large variety of image-related tasks.

\end{abstract}

\section{Introduction}
In the past years, deep neural networks such as convolutional or recurrent ones
have become highly popular for solving various prediction
problems, notably in computer vision and natural language processing. Conceptually close
to approaches that were developed several decades ago~(see,
\cite{lecun-98x}), they greatly benefit from
the large amounts of labeled data that have been made available recently, allowing
to learn huge numbers of model parameters without worrying too much about
overfitting. Among other reasons explaining their success, the
engineering effort of the deep learning community and various methodological
improvements have made it possible to learn in a day on a GPU complex models that
would have required weeks of computations on a traditional
CPU~(see, \eg, \cite{he2015deep,krizhevsky2012,simonyan2014very}).

Before the resurgence of neural networks, non-parametric models based on
positive definite kernels were one of the most dominant topics in machine
learning~\cite{scholkopf2002learning}. These approaches are still
widely used today because of several attractive features. Kernel
methods are indeed versatile; as long as a positive definite kernel is specified for the
type of data considered---\emph{e.g.}, vectors, sequences, graphs, or sets---a large class of machine learning algorithms
originally defined for linear models may be used. This family include supervised
formulations such as support vector machines and unsupervised ones such
as principal or canonical component analysis, or K-means and spectral clustering.
\emph{The problem of data representation is thus decoupled from that of
learning theory and algorithms.} Kernel methods also admit natural mechanisms to
control the learning capacity and reduce overfitting~\cite{scholkopf2002learning}.

On the other hand, traditional kernel methods suffer from several drawbacks.
The first one is their computational complexity, which grows quadratically with the sample size due to the
computation of the Gram matrix. Fortunately, significant progress has been
achieved to solve the scalability issue, either by exploiting low-rank
approximations of the kernel matrix~\cite{williams2001,zhang2008improved}, or with random sampling techniques for shift-invariant
kernels~\cite{rahimi2007}. The second
disadvantage is more critical; by decoupling learning and data representation,
kernel methods seem by nature incompatible with end-to-end learning---that is,
the representation of data adapted to the task at hand, which is the
cornerstone of deep neural networks and one of the main reason of their
success. \emph{The main objective of this paper is precisely to tackle this
issue in the context of image modeling.}

Specifically, our approach is based on convolutional kernel
networks, which have been recently introduced in~\cite{mairal2014convolutional}. Similar to hierarchical kernel descriptors~\cite{bo2011}, local image
neighborhoods are mapped to points in a reproducing kernel Hilbert space
via the kernel trick. Then, hierarchical representations are built via kernel
compositions, producing a sequence of ``feature maps'' akin to
convolutional neural networks, but of infinite dimension. 
To make the image
model computationally tractable, convolutional kernel networks provide an approximation
scheme that can be interpreted as a particular type of convolutional neural
network learned without supervision. 

To perform end-to-end learning given labeled data, \emph{we use a simple but effective
principle consisting of learning discriminative subspaces in RKHSs, where
we project data}. We implement this idea in the context of convolutional kernel
networks, where linear subspaces, one per layer, are jointly optimized by
minimizing a supervised loss function. The formulation turns out to
be a new type of convolutional neural network with a non-standard
parametrization. 
The network also admits simple principles to
learn without supervision: learning the subspaces may be indeed
achieved efficiently with classical kernel approximation
techniques~\cite{williams2001,zhang2008improved}.

To demonstrate the effectiveness of our approach in various contexts, we
consider image classification benchmarks such as
CIFAR-10~\cite{krizhevsky2012} and SVHN~\cite{netzer2011reading}, which are
often used to evaluate deep neural networks; then, we adapt our model to perform 
image super-resolution, which is a challenging inverse problem.
On the SVHN and CIFAR-10 datasets, we obtain a competitive accuracy, with about $2\%$ and $10\%$ error rates,
respectively, without model averaging or data augmentation. For image
up-scaling, we
outperform recent approaches based on classical convolutional neural
networks~\cite{dong2014learning,dong2015image}. 

We believe that these results are highly promising. Our image model achieves competitive performance
in two different contexts, paving the
way to many other applications. Moreover, our results are also subject to
improvements. In particular, we did not use GPUs yet, which has limited our ability to exhaustively explore model
hyper-parameters and evaluate the accuracy of large networks. We also did
not investigate classical regularization/optimization techniques such as
Dropout~\cite{krizhevsky2012}, batch normalization~\cite{ioffe2015batch}, or
recent advances allowing to train very deep
networks~\cite{he2015deep,simonyan2014very}. 
To gain more scalability and start exploring these directions, we are currently working on a GPU implementation, which we plan to publicly release along with our current CPU implementation.

\paragraph{Related Deep and Shallow Kernel Machines.}
One of our goals is to make a bridge between kernel methods and deep networks,
and ideally reach the best of both worlds. Given the potentially attractive features of
such a combination, several attempts have been made in the past to
unify these two schools of thought.  A first proof of
concept was introduced in~\cite{cho2009} with
the arc-cosine kernel, which admits an integral representation that can be
interpreted as a one-layer neural network with random weights and infinite
number of rectified linear units. Besides, a multilayer kernel may be obtained by
kernel compositions~\cite{cho2009}. Then, hierarchical kernel
descriptors~\cite{bo2011} and convolutional kernel
networks~\cite{mairal2014convolutional} extend a similar idea in the context of
images leading to unsupervised representations~\cite{mairal2014convolutional}.

Multiple kernel learning ~\cite{sonnenburg2006large} is also related to our
work since is it is a notable attempt to introduce supervision in the
kernel design. It provides techniques to select a combination of kernels
from a pre-defined collection, and typically requires to have already ``good''
kernels in the collection to perform well. More related to our work, the
backpropagation algorithm for the Fisher kernel introduced
in~\cite{sydorov2014deep} learns the parameters of a Gaussian mixture model
with supervision. In comparison, our approach does not require a probabilistic
model and learns parameters at several layers. 
Finally, we note that a concurrent effort to ours is conducted in the Bayesian
community with deep Gaussian processes~\cite{damianou}, complementing the
Frequentist approach that we follow in our paper.

\section{Learning Hierarchies of Subspaces with Convolutional Kernel Networks}\label{sec:unsup}
\begin{figure}
   \centering
   \includegraphics[width=0.9\linewidth]{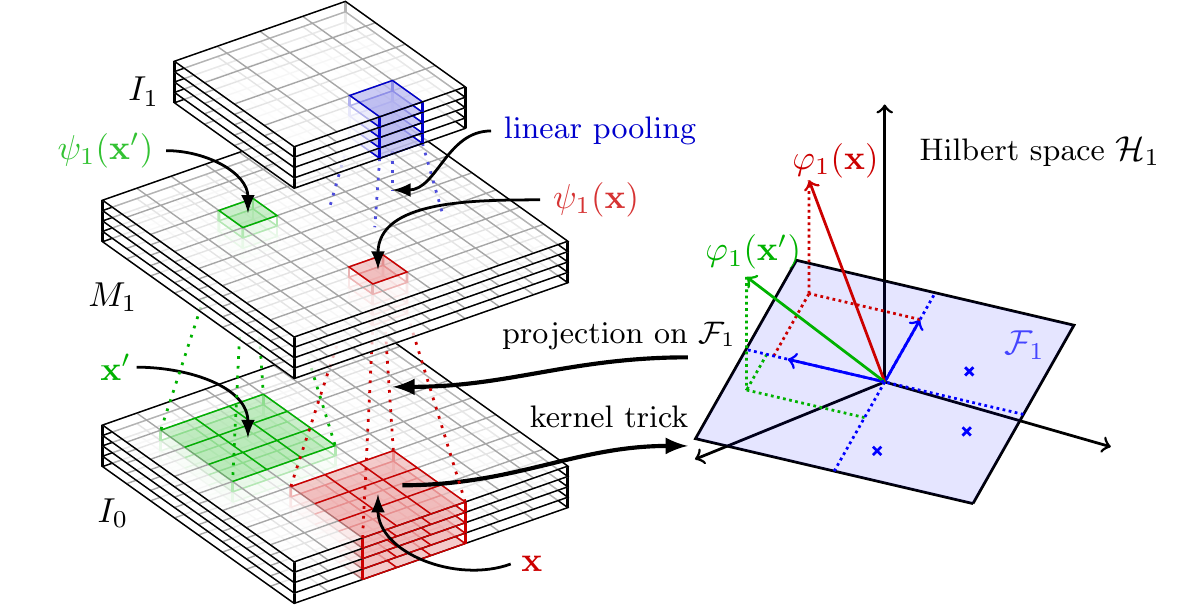}
   \caption{Our variant of convolutional kernel networks, illustrated between layers~$0$ and~$1$. 
      Local patches (receptive fields) are mapped to the RKHS
      $\Hcal_1$ via the kernel trick and then projected to the
      finite-dimensional subspace~$\Fcal_1 \!=\! \text{Span}(\varphi(\z_1),\ldots,\varphi(\z_{p_1}))$. The small blue crosses on the right represent the points $\varphi(\z_1),\ldots,\varphi(\z_{p_1})$. With no supervision, optimizing
      $\Fcal_1$ consists of minimizing projection residuals. With supervision, the
      subspace is optimized via back-propagation.  Going from layer~$k$ to
   layer~$k+1$ is achieved by stacking the model described here and shifting
indices.}\label{fig:ckn}
\end{figure}

In this section, we present the principles of convolutional kernel
networks and a few generalizations and
improvements of the original approach of~\cite{mairal2014convolutional}. Essentially, the model builds upon four
ideas that are detailed below and that are illustrated in Figure~\ref{fig:ckn} for a model with a single layer.

\paragraph{Idea 1: use the kernel trick to represent local image neighborhoods in a RKHS.}~\\
Given a set~$\Xcal$, a positive definite kernel
$K: \Xcal \times \Xcal \to \Real$ implicitly
defines a Hilbert space~$\Hcal$, called
reproducing kernel Hilbert space (RKHS), along with a mapping~$\varphi: \Xcal
\to \Hcal$. This embedding is such that the kernel value $K(\x,\x')$
corresponds to the inner product $\langle \varphi(\x),\varphi(\x')\rangle_{\Hcal}$.
Called ``kernel trick'', this approach can be used to obtain nonlinear representations of local
image patches~\cite{bo2011,mairal2014convolutional}.

More precisely, consider an image~$I_0: \Omega_0 \to \Real^{p_0}$, where $p_0$ is the
number of channels, \eg, $p_0=3$ for RGB, and $\Omega_0 \subset [0,1]^2$ is a set of pixel
coordinates, typically a two-dimensional grid. Given two image patches~$\x,\x'$ of size $e_0 \times e_0$, 
represented as vectors in~$\Real^{p_0 e_0^2}$, we define a kernel~$K_1$ as
\begin{equation}
   \textstyle K_1(\x,\x') =  \|\x\|\, \|\x'\|\,  \kappa_1\left( \left\langle \frac{\x}{\|\x\|} , \frac{\x'}{\|\x'\|} \right\rangle \right) ~~\text{if}~~ \x,\x' \neq 0~~~\text{and}~0~\text{otherwise},  \label{eq:kernel}
\end{equation}
where $\|.\|$ and~$\langle.,.\rangle$ denote the usual Euclidean norm and
inner-product, respectively, and $\kappa_1(\langle .,.\rangle)$ is a
dot-product kernel on the sphere. Specifically, 
$\kappa_1$ should be smooth and its Taylor expansion have non-negative
coefficients to ensure positive definiteness~\cite{scholkopf2002learning}. For example, the arc-cosine~\cite{cho2009} or the
Gaussian (RBF) kernels may be used: given two vectors
$\y,\y'$ with unit $\ell_2$-norm, choose for instance
\begin{equation}
   \kappa_1(\langle \y,\y' \rangle) =  e^{\alpha_1 \left(\langle \y,\y' \rangle -1 \right)} =  e^{-\frac{\alpha_1}{2} \|\y-\y'\|_2^2}. \label{eq:RBF}
\end{equation}
Then, we have implicitly defined the RKHS $\Hcal_1$ associated to~$K_1$ and a
mapping $\varphi_1:\Real^{p_0 e_0^2} \to \Hcal_1$.

\paragraph{Idea 2: project onto a finite-dimensional subspace of the RKHS with convolution layers.} ~\\
The representation of patches in a RKHS requires
finite-dimensional approximations to be computationally manageable.
The original model of~\cite{mairal2014convolutional} does that by exploiting an integral form
of the RBF kernel. Specifically, given two patches $\x$ and $\x'$, convolutional kernel
networks provide two vectors $\psi_1(\x),\psi_1(\x')$ in~$\Real^{p_1}$ such
that  the kernel value $\langle\varphi_1(\x), \varphi_1(\x') \rangle_{\Hcal_1}$
is close to the Euclidean inner product $\langle \psi_1(\x), \psi_1(\x')
\rangle$.
After applying this transformation to all overlapping patches of the input image~$I_0$, a
spatial map $M_1: \Omega_0 \to \Real^{p_1}$ may be obtained such that for all~$z$ in~$\Omega_0$,
$M_1(z) = \psi_1(\x_z)$, where $\x_z$ is the $e_0 \times e_0$ patch from $I_0$
centered at pixel location $z$.\footnote{To simplify, we use zero-padding when patches are close to the image boundaries, but this is optional.}
With the approximation scheme
of~\cite{mairal2014convolutional}, $M_1$ can be interpreted as the output feature map of a
one-layer convolutional neural network.   
 
A conceptual drawback of~\cite{mairal2014convolutional} is that data points $\varphi_1(\x_1),\varphi_1(\x_2),\ldots$ are approximated
by vectors that do not live in the RKHS~$\Hcal_1$.
This issue can be solved by using variants of the
Nystr\"om method~\cite{williams2001}, which consists of projecting data
onto a subspace
of~$\Hcal_1$ with finite dimension~$p_1$. 
For this task, we have adapted the approach of~\cite{zhang2008improved}: we
build a database of $n$ patches $\x_1,\ldots,\x_n$ randomly
extracted from various images and normalized to have unit $\ell_2$-norm, and perform a spherical $K$-means algorithm to obtain $p_1$
centroids $\z_1,\ldots,\z_{p_1}$ with unit $\ell_2$-norm. Then, a new patch~$\x$ is approximated by its projection
onto the $p_1$-dimensional subspace $\Fcal_1 \!=\! \text{Span}(\varphi(\z_1),\ldots,\varphi(\z_{p_1}))$.

The projection of~$\varphi_1(\x)$ onto~$\Fcal_1$ admits a natural
parametrization $\psi_1(\x)$ in~$\Real^{p_1}$ .
The explicit formula is classical (see~\cite{williams2001,zhang2008improved}
and Appendix~\ref{appendix:A}), leading to
\begin{equation}
   \psi_1(\x) \defin  \|\x\| \kappa_1(\Z^\top \Z)^{-1/2} \kappa_1\left(\Z^\top\frac{\x}{\|\x\|}\right)  ~~\text{if}~~\x \neq 0~~\text{and}~~0~~\text{otherwise}, \label{eq:psi}
\end{equation}
where we have introduced the matrix $\Z=[\z_1,\ldots,\z_{p_1}]$,
and, by an abuse of notation, the function $\kappa_1$ is applied pointwise to
its arguments. Then, the spatial map $M_1: \Omega_0
\to \Real^{p_1}$ introduced above can be obtained by (i) computing the quantities $\Z^\top \x$ for all patches $\x$ of the image~$I$ (spatial convolution after mirroring the filters $\z_j$);
(ii) contrast-normalization involving the norm $\|\x\|$; (iii) applying the pointwise non-linear function~$\kappa_1$; (iv) applying the linear transform 
$\kappa_1(\Z^\top \Z)^{-1/2}$ at every pixel location (which may be seen as $1 \times 1$ spatial convolution); (v) multiplying by the norm $\|\x\|$ making $\psi_1$ homogeneous.
In other words, we obtain a particular convolutional neural network, with non-standard parametrization.
Note that learning requires only performing a K-means algorithm and computing
the inverse square-root matrix~$\kappa_1(\Z^\top \Z)^{-1/2}$; therefore, the
training procedure is very fast.

Then, it is worth noting that the encoding function~$\psi_1$ with
kernel~(\ref{eq:RBF}) is reminiscent of radial basis function networks
(RBFNs)~\cite{broomhead1988radial}, whose hidden layer resembles~(\ref{eq:psi})
without the matrix~$\kappa_1(\Z^\top \Z)^{-1/2}$ and with no normalization.
The difference between RBFNs and our model is nevertheless significant. The RKHS mapping,
which is absent from RBFNs, is indeed a key to the multilayer construction that will be presented shortly:
a network layer takes points from the RKHS's previous layer as input and use
the corresponding RKHS inner-product. To the best of our knowledge, there is no similar
multilayer and/or convolutional construction in the radial basis function network literature. 

\paragraph{Idea 3: linear pooling in~$\Fcal_1$ is equivalent to linear pooling on the finite-dimensional map~$M_1$.}~\newline
The previous steps transform an
image $I_0: \Omega_0 \to \Real^{p_0}$ into a map $M_1: \Omega_0 \to
\Real^{p_1}$, where each vector $M_1(z)$ in~$\Real^{p_1}$ encodes a point in~$\Fcal_1$
representing information of a local image neighborhood centered at location $z$.
Then, convolutional kernel networks involve a pooling step to gain
invariance to small shifts, leading to another finite-dimensional map $I_1:
\Omega_1 \to \Real^{p_1}$ with smaller resolution:
\begin{equation}
   I_1(z) =  \sum_{z' \in \Omega_0} M_1(z') e^{-\beta_1 \|z'-z\|_2^2}.
\end{equation}
The Gaussian weights act as an anti-aliasing filter
for downsampling the map~$M_1$ and~$\beta_1$ is set according to the desired
subsampling factor~(see \cite{mairal2014convolutional}), which does not need to
be integer. 
Then, every point $I_1(z)$ in~$\Real^{p_1}$ may be interpreted as a linear combination
of points in~$\Fcal_1$, which is itself in~$\Fcal_1$ since $\Fcal_1$ is a
linear subspace. Note that the linear pooling step was originally motivated
in~\cite{mairal2014convolutional} as an approximation scheme for a match
kernel, but this point of view is not critically important here.

\paragraph{Idea 4: build a multilayer image representation by stacking and composing kernels.}~\newline
By following the first three principles described above, the input image~$I_0: \Omega_0 \to
\Real^{p_0}$ is transformed into another one $I_1: \Omega_1 \to \Real^{p_1}$.
It is then straightforward to apply again the same procedure to obtain another
map $I_2: \Omega_2 \to \Real^{p_2}$, then $I_3: \Omega_3 \to \Real^{p_3}$, \emph{etc}. 
By going up in the hierarchy, the vectors $I_k(z)$ in~$\Real^{p_k}$ represent larger and larger
image neighborhoods (aka. receptive fields) with more invariance gained by the pooling layers,
akin to classical convolutional neural networks.

The multilayer scheme produces a sequence of maps $(I_k)_{k \geq 0}$, where
each vector $I_k(z)$ encodes a point---say $f_k(z)$---in the linear subspace
$\Fcal_k$ of~$\Hcal_{k}$. Thus, we implicitly represent an image at
layer~$k$ as a spatial map $f_k: \Omega_k \to \Hcal_k$ such that
$\langle I_k(z), I_k'(z') \rangle = \langle f_k(z), f_k'(z')
\rangle_{\Hcal_k}$ for all~$z,z'$.
As mentioned previously, the mapping to the RKHS is a key to the multilayer
construction. Given~$I_k$, larger image neighborhoods are represented
by patches of size $e_k \times e_k$ that can be mapped to a point in the Cartesian product space
$\Hcal_k^{e_k \times e_k}$ endowed with its natural inner-product; 
finally, the kernel~$K_{k+1}$ defined on these patches can be seen as a
kernel on larger image neighborhoods than~$K_k$.

\section{End-to-End Kernel Learning with Supervised CKNs}
In the previous section, we have described a variant of convolutional kernel
networks where linear subspaces are learned at every layer. This is achieved without supervision 
by a K-means algorithm leading to small projection residuals.
It is thus natural to introduce also a discriminative approach. 

\subsection{Backpropagation Rules for Convolutional Kernel Networks}
We now consider a prediction task, where we are given a training set of images
$I_0^1,I_0^2,\ldots,I_0^n$ with respective scalar labels $y_1,\ldots,y_n$ living either 
in $\{-1; +1\}$ for binary classification and~$\Real$ for regression. 
For simplicity, we only present these two settings here, but extensions to
multiclass classification and multivariate regression are straightforward.
We also assume that we are given a smooth convex loss function $L: \Real \times \Real \to \Real$ that measures
the fit of a prediction to the true label $y$. 

Given a positive definite kernel $K$ on images, the classical empirical risk
minimization formulation consists of finding a prediction function in the RKHS~$\Hcal$ associated to~$K$ 
by minimizing the objective
\begin{equation}
   \min_{f \in \Hcal} \frac{1}{n} \sum_{i=1}^n L( y_i, f(I_0^i))  + \frac{\lambda}{2} \| f \|_{\Hcal}^2,\label{eq:erm}
\end{equation}
where the parameter~$\lambda$ controls the smoothness of the prediction
function~$f$ with respect to the geometry induced by the
kernel, hence regularizing and reducing
overfitting~\cite{scholkopf2002learning}.
After training a convolutional kernel network with $k$ layers, such a positive definite kernel may be defined as 
\begin{equation}
K_\Zcal(I_0,I_0')=\sum_{z \in \Omega_k} \langle f_k(z),f_k'(z) \rangle_{\Hcal_k}=\sum_{z \in \Omega_k} \langle I_k(z),I_k'(z) \rangle, 
\end{equation}
where $I_k,I_k'$ are the $k$-th finite-dimensional feature maps of $I_0,I_0'$, respectively, and $f_k, f_k'$ the corresponding maps in $\Omega_k \to \Hcal_k$,
which have been defined in the previous section. The kernel is also indexed by~$\Zcal$, which represents the network parameters---that is, the subspaces $\Fcal_1,\ldots,\Fcal_k$, or equivalently the set of filters $\Z_1,\ldots,\Z_k$ from Eq.~(\ref{eq:psi}).
Then, formulation~(\ref{eq:erm}) becomes equivalent to 
\begin{equation}
   \min_{\W \in \Real^{p_k \times |\Omega_k|}} \frac{1}{n} \sum_{i=1}^n L( y_i, \langle \W , I_k^i \rangle )  + \frac{\lambda}{2} \| \W \|_{\text{F}}^2, \label{eq:erm2}
\end{equation}
where $\|.\|_{\text{F}}$ is the Frobenius norm that extends the Euclidean norm
to matrices, and, with an abuse of notation, the maps $I_k^i$ are seen as
matrices in $\Real^{p_k \times |\Omega_k|}$.
Then, \emph{the supervised convolutional kernel network formulation consists of jointly
minimizing~(\ref{eq:erm2}) with respect to $\W$ in $\Real^{p_k \times
|\Omega_k|}$ and with respect to the set of filters $\Z_1,\ldots,\Z_k$},
whose columns are constrained to be on the Euclidean sphere.

\paragraph{Computing the derivative with respect to the filters $\Z_1,\ldots,\Z_k$.} ~\\
Since we consider a smooth loss function~$L$, \eg, logistic, squared hinge, or
square loss, optimizing~(\ref{eq:erm2}) with respect to~$\W$ can be achieved
with any gradient-based method. Moreover, when~$L$ is convex, we may also use
fast dedicated solvers, (see, \eg,~\cite{hongzhou}, and references therein).
Optimizing with respect to the filters~$\Z_j$, $j=1,\ldots,k$ is more involved
because of the lack of convexity. Yet, the objective function is differentiable,
and there is hope to find a ``good'' stationary point by using classical 
stochastic optimization techniques that have been successful for training 
deep networks.

For that, we need to compute the gradient by using the chain rule---also called
``backpropagation''~\cite{lecun-98x}. We instantiate this rule in the
next lemma, which we have found useful to simplify the calculation.
\vsp
\begin{lemma}[Perturbation view of backpropagration.]~\label{lemma:backprop}\newline
Consider an image~$I_0$ represented here as a matrix in~$\Real^{p_0 \times
|\Omega_0|}$, associated to a label~$y$ in~$\Real$ and call~$I_{k}^{\Zcal}$ the
$k$-th feature map obtained by encoding $I_0$ with the network parameters~$\Zcal$. Then,
consider a perturbation $\Ecal = \{ \varepsilonb_1, \ldots, \varepsilonb_k \}$  
of the set of filters~$\Zcal$. Assume that we have for all~$j\geq 0$,
\begin{equation}
   I_{j}^{\Zcal+\Ecal} = I_{j}^{\Zcal} + \Delta I_{j}^{\Zcal,\Ecal} + o(\|\Ecal\|), \label{eq:perturb}
\end{equation}
where $\|\Ecal\|$ is equal to $\sum_{l=1}^k\|\varepsilonb_l\|_{\text{F}}$, and $\Delta I_{j}^{\Zcal,\Ecal}$ is a matrix in~$\Real^{p_j \times |\Omega_j|}$ such that for all matrices~$\U$ of the same size,
\begin{equation}
   \langle \Delta I_{j}^{\Zcal,\Ecal}, \U \rangle = \langle \varepsilonb_j, g_j(\U) \rangle + \langle  \Delta I_{j-1}^{\Zcal,\Ecal}, h_j(\U) \rangle,  \label{eq:perturb2}
\end{equation}
where the inner-product is the Frobenius's one and $g_j, h_j$ are linear functions.
Then, 
\begin{equation}
   \nabla_{\Z_j} L(y,  \langle \W , I_k^\Zcal \rangle ) = L'(y,  \langle \W , I_k^\Zcal \rangle ) \, g_j(h_{j+1}(\ldots h_k(\W)),\label{eq:gradient}
\end{equation}
where $L'$ denote the derivative of the smooth function $L$ with respect to its second argument.
\end{lemma}
The proof of this lemma is straightforward and follows from the
definition of the Fr\'echet derivative. Nevertheless, it is useful
to derive the closed form of the gradient in the next proposition. 
\vsp
\begin{proposition}[Gradient of the loss with respect to the the filters~$\Z_1,\ldots,\Z_k$.]~\label{prop:gradient}\newline
   Consider the quantities introduced in Lemma~\ref{lemma:backprop}, but denote $I_j^\Zcal$ by~$I_j$ for simplicity. By construction, we have for all $j \geq 1$,
   \begin{equation}
      I_j =   \A_j \kappa_j( \Z_j^\top \E_{j}(I_{\jmone}) \S_j^{-1}) \S_j \P_j,  \label{eq:forward}
   \end{equation}
   where $I_{j}$ is seen as a matrix in $\Real^{p_{j}
   \times |\Omega_{j}|}$; $\E_j$ is the linear operator that
   extracts all overlapping $e_{\jmone} \times e_{\jmone}$ patches
   from a map such that $\E_{j}(I_{\jmone})$ is a matrix of
   size $p_{\jmone}e_{\jmone}^2 \times |\Omega_{\jmone}|$; $\S_j$ is a diagonal matrix
   whose diagonal entries carry the $\ell_2$-norm of the columns of
   $\E_{j}(I_\jmone)$; $\A_j$ is short for $\kappa_j(\Z_j^\top
   \Z_j)^{-1/2}$; and~$\P_j$ is a matrix of size $|\Omega_{\jmone}| \times |\Omega_j|$ performing the linear pooling operation.
   Then, the gradient of the loss with respect to the filters $\Z_j$,
   $j=1,\ldots,k$ is given by~(\ref{eq:gradient}) with
   \begin{equation}
      \begin{split}
         g_j(\U) & = \E_j(I_{\jmone}) \B_j^\top  - \frac{1}{2} \Z_j \left( \kappa_j'(\Z_j^\top \Z_j) \odot (\C_j + \C_j^\top) \right)   \\
         h_j(\U) & = \E_j^\star \left(  \Z_j \B_j  +  \E_j(I_{\jmone}) \left(\S_j^{-2} \odot \left( M_j^{\top} \U \P_j^\top -  \E_j(I_{\jmone})^\top \Z_j\B_j\right)\right)  \right),
      \end{split}\label{eq:gjhj}
   \end{equation}
   where~$\U$ is any matrix of the same size as~$I_j$, $M_j= \A_j \kappa_j( \Z_j^\top \E_{j}(I_{\jmone}) \S_j^{-1}) \S_j$ is the $j$-th feature map before the pooling step, $\odot$ is the Hadamart (elementwise) product, $\E_j^\star$ is the adjoint of~$\E_j$, 
   and 
   \begin{equation}
      \B_j = \kappa_j'\left(\Z_j^\top \E_j(I_{\jmone}) \S_j^{-1}\right) \odot \left(\A_j \U \P_j^\top\right)   ~~~~\text{and}~~~~ \C_j =  \A_j^{1/2} I_j\U^\top \A_j^{3/2}.
   \end{equation}
\end{proposition}
The proof is presented in Appendix~\ref{appendix:proof}.
Most quantities that appear above admit 
physical interpretations: multiplication by~$\P_j$ performs 
downsampling; multiplication by~$\P_j^\top$
performs upsampling; multiplication of~$\E_j(I_{\jmone})$ on the right by~$\S_j^{-1}$ performs
$\ell_2$-normalization of the columns;
$\Z_j^\top\E_j(I_{\jmone})$ can be seen as a spatial convolution of the
map~$I_\jmone$ by the filters~$\Z_j$; finally, $\E_j^\star$ ``combines'' a
set of patches into a spatial map by adding to each pixel location the
respective patch contributions.

Computing the gradient 
requires a forward pass to obtain the maps~$I_j$ through~(\ref{eq:forward}) and a backward
pass that composes the functions $g_j, h_j$ as in~(\ref{eq:gradient}). The
complexity of the forward step is dominated by the convolutions
$\Z_j^\top\E_j(I_{\jmone})$, as in convolutional neural
networks. 
The cost of the backward pass 
is the same as the forward one up to a constant factor. 
Assuming $p_j \!\leq\! |\Omega_{\jmone}|$, which is typical for 
lower layers that require more computation than upper ones, 
the most expensive cost is due to 
$\E_j(I_{\jmone})\B_j^\top$ and $\Z_j\B_j$ which is the same as $\Z_j^\top\E_j(I_{\jmone})$. We also pre-compute $\A_j^{1/2}$ and~$\A_j^{3/2}$ by eigenvalue decompositions, whose
cost is reasonable when performed only once per minibatch.
Off-diagonal elements of $M_j^{\top} \U\P_j^\top -
\E_j(I_{\jmone})^\top \Z_j\B_j$ are also not computed since they are set to zero after elementwise multiplication with a diagonal matrix.
 In practice, we also replace $\A_j$
 by $(\kappa_j(\Z_j^\top \Z_j) + \varepsilon\I)^{-1/2}$ with
$\varepsilon\!=\!0.001$, which corresponds to performing a regularized
projection onto~$\Fcal_j$ (see Appendix~\ref{appendix:A}). Finally, 
a small offset of $0.00001$ is added to the diagonal entries of $\S_j$.

\vsp
\paragraph{Optimizing hyper-parameters for RBF kernels.}
When using the kernel~(\ref{eq:RBF}), the objective is differentiable
with respect to the hyper-parameters~$\alpha_j$. 
When large amounts of training data are available and overfitting is not a
issue, optimizing the training loss by taking 
gradient steps with respect to these parameters seems appropriate instead of using a canonical parameter value.
Otherwise, more involved techniques may be needed; 
we plan to investigate other strategies in future work.

\subsection{Optimization and Practical Heuristics}
The backpropagation rules of the previous section have set up the stage for using a
stochastic gradient descent method (SGD). We now present a few strategies to accelerate it in our context.

\vsp
\paragraph{Hybrid convex/non-convex optimization.}
Recently, many incremental optimization techniques have been proposed for solving
\emph{convex} optimization problems of the form~(\ref{eq:erm2}) when $n$ is
large but finite (see \cite{hongzhou} and references therein). These methods
usually provide a great speed-up over the stochastic gradient descent algorithm
without suffering from the burden of choosing a learning
rate. The price to pay is that they rely on convexity, and they require
storing into memory the full training set. For solving~(\ref{eq:erm2}) with \emph{fixed} network parameters~$\Zcal$, it means storing
the $n$ maps~$I_k^i$, which is often reasonable if we do not use data
augmentation. To partially leverage these fast algorithms for our non-convex problem, we have adopted 
a minimization scheme that alternates between two steps: (i) fix $\Zcal$, then make a forward pass on the data to
compute the $n$ maps~$I_k^i$ and minimize the convex problem~(\ref{eq:erm2}) with respect to~$\W$ using the accelerated
MISO algorithm~\cite{hongzhou}; (ii) fix~$\W$, then make one pass of a projected
stochastic gradient algorithm to update the $k$ set of filters~$\Z_j$.  The set of
network parameters~$\Zcal$ is initialized with the unsupervised learning
method described in Section~\ref{sec:unsup}.

\paragraph{Preconditioning on the sphere.}
The kernels~$\kappa_j$ are defined on the
sphere; therefore, it is natural to constrain the filters---that is, the columns of the
matrices~$\Z_j$---to have unit~$\ell_2$-norm. As a result, a classical
stochastic gradient descent algorithm updates at iteration~$t$ each filter~$\z$
as follows $\z \leftarrow \text{Proj}_{\|.\|_2= 1}[ \z - \eta_t \nabla_\z
L_t]$, where $\nabla_\z L_t$ is an estimate of the gradient
computed on a minibatch and~$\eta_t$ is a learning rate.
In practice, we found that convergence could be accelerated by preconditioning,
which consists of optimizing after a change of variable to reduce the correlation of
gradient entries. 
For \emph{unconstrained} optimization,  this heuristic involves choosing a
symmetric positive definite matrix~$\Q$ and replacing the update direction
$\nabla_\z L_t$ by $\Q\nabla_\z L_t$, or, equivalently, performing the change
of variable $\z =\Q^{1/2}\z'$ and optimizing over~$\z'$. 
When constraints are present, the case is not as simple since $\Q\nabla_\z
L_t$ may not be a descent direction.
Fortunately, it is possible to exploit the manifold structure of the constraint set
(here, the sphere) to perform an appropriate update~\cite{absil2009optimization}.
Concretely, (i) we choose a matrix~$\Q$ per layer that is equal to the inverse
covariance matrix of the patches from the same layer computed after the
initialization of the network parameters. (ii) We perform stochastic gradient
descent steps on the sphere manifold after the change of variable $\z
=\Q^{1/2}\z'$, leading to the update
$\z \leftarrow \proj[\z - \eta_t (\I - (1/{\z^\top \Q \z}){\Q \z \z^\top}) \Q\nabla_\z L_t]$.
Because this heuristic is not a critical component, but simply an improvement of SGD,
we relegate mathematical details in
Appendix~\ref{appendix:precond}.

\vsp
\paragraph{Automatic learning rate tuning.}
Choosing the right learning rate in stochastic optimization is still an
important issue despite the large amount of work existing on the topic, see,
\eg, \cite{lecun-98x} and references therein. 
In our paper, we use the following basic heuristic:
the initial learning rate~$\eta_t$ is chosen ``large
enough''; then, the training loss is evaluated after each update of the
weights~$\W$. 
When the training loss increases between two epochs, we simply divide the
learning rate by two, and perform ``back-tracking'' by replacing the current
network parameters by the previous ones.

\vsp
\paragraph{Active-set heuristic.}
For classification tasks, ``easy'' samples have often
negligible contribution to the gradient (see, \eg,~\cite{lecun-98x}). For instance, for the
squared hinge loss $L(y,\hat{y})=\max(0,1-y \hat{y})^2$, the gradient vanishes when the margin
$y \hat{y}$ is greater than one. 
This motivates the following heuristic: we consider a set of
active samples, initially all of them, and remove a sample from the active
set as soon as we obtain zero when computing its gradient. 
In the subsequent optimization steps, only active samples are considered, and after each epoch,
we randomly reactivate $10\%$ of the inactive ones.

\section{Experiments}
We now present experiments on image classification and super-resolution. 
All experiments were conducted on $8$-core and~$10$-core $2.4$GHz Intel CPUs using C++ and Matlab.

\subsection{Image Classification on ``Deep Learning'' Benchmarks}\label{subsec:class}
We consider the datasets CIFAR-10~\cite{krizhevsky2012} and
SVHN~\cite{netzer2011reading}, 
which contain $32\times 32$ images from~$10$
classes. CIFAR-10 is medium-sized with $50\,000$ training samples and~$10\,000$
test ones. SVHN is larger with $604\,388$ training examples and $26\,032$ test
ones. We evaluate the performance of a $9$-layer network, designed with few
hyper-parameters: for each layer, we learn $512$ filters and choose the
RBF kernels~$\kappa_j$ defined in~(\ref{eq:RBF}) with initial parameters~$\alpha_j\!=\!1/(0.5^2)$.
Layers~$1,3,5,7,9$ use $3 \!\times\! 3$ patches and a subsampling pooling
factor of~$\sqrt{2}$ except for layer~$9$ where the factor is $3$; Layers~$2,4,6,8$ use
simply~$1 \times 1$ patches and no subsampling. For CIFAR-10, the
parameters~$\alpha_j$ are kept fixed during training, and for SVHN, they
are updated in the same way as the filters.
We use the squared hinge loss in a one vs all setting to perform multi-class
classification (with shared filters~$\Zcal$ between
classes). The input of the network is pre-processed with the local whitening procedure
described in~\cite{paulin2015local}.
We use the optimization heuristics from the previous section, notably the automatic learning rate scheme, and
a gradient momentum with parameter~$0.9$, following~\cite{krizhevsky2012}. The regularization parameter~$\lambda$ and
the number of epochs are set by first running the algorithm on a $80/20$ validation split 
of the training set. $\lambda$ is chosen near the canonical parameter~$\lambda=1/n$, in
the range~$2^i/n$, with $i=-4,\ldots,4$, and the number of epochs is at
most~$100$. The initial learning rate is~$10$ with a minibatch size of~$128$.

We present our results in Table~\ref{table:class} along with the performance
achieved by a few recent methods without data augmentation or model
voting/averaging. In this context, the best published results
are obtained by the generalized pooling scheme
of~\cite{lee2016generalizing}.
We achieve about $2\%$ test error on SVHN and about~$10\%$ on CIFAR-10, which positions our method 
as a reasonably ``competitive'' one, in the same ballpark as the deeply
supervised nets of~\cite{lee2015} or network in network of~\cite{lin2013network}.

\begin{table}[hbtp]
   \caption{Test error in percents reported by a few recent publications on the CIFAR-10 and SVHN datasets~without data augmentation or model voting/averaging.}\label{table:class}
   \vspace*{-0.1cm}
   \centering
   \renewcommand\tabcolsep{0.16cm}
   \footnotesize
   \begin{tabular}{|c|*{5}{>{\centering}m{1.63cm}|}|c|   }
      \hline
      & Stoch P.~\cite{zeiler2013} & MaxOut~\cite{goodfellow2013} & NiN~\cite{lin2013network} & DSN~\cite{lee2015} & Gen P. \cite{lee2016generalizing}  & SCKN (Ours) \\
      \hline
      CIFAR-10 & 15.13 & 11.68 & 10.41 & 9.69 & \textbf{7.62} & 10.20 \\
      \hline
      SVHN  & 2.80 & 2.47 & 2.35 & 1.92 & \textbf{1.69} & 2.04 \\
      \hline
   \end{tabular}
   \vspace*{-0.2cm}
\end{table}

Due to lack of space, the results reported here only include a single
supervised model. Preliminary experiments with no supervision show also
that one may obtain competitive accuracy with wide shallow architectures. For
instance, a two-layer network with (1024-16384) filters achieves $14.2\%$ error
on CIFAR-10. Note also that our unsupervised model outperforms 
original CKNs~\cite{mairal2014convolutional}. The best single model
from~\cite{mairal2014convolutional} gives indeed $21.7\%$.
Training the same architecture with our approach is two orders of magnitude
faster and gives $19.3\%$. Another aspect we did not study is model complexity.
Here as well, preliminary experiments are encouraging. Reducing the number of
filters to~$128$ per layer yields indeed $11.95\%$ error on CIFAR-10 and $2.15\%$
on SVHN. A more precise comparison with no supervision and with various
network complexities will be presented in another venue.

\subsection{Image Super-Resolution from a Single Image}
Image up-scaling is a challenging
problem, where convolutional neural networks have
obtained significant success~\cite{dong2014learning,dong2015image,wang2015deep}.
Here, we follow~\cite{dong2015image} and replace traditional convolutional neural networks by our
supervised kernel machine. Specifically, RGB images are converted to the YCbCr
color space and the upscaling method is applied to the luminance channel only to make the comparison possible with previous work.  Then, the problem
is formulated as a multivariate regression one. We build a
database of $200\,000$ patches of size $32 \times 32$ randomly extracted from the
BSD500 dataset~\cite{arbelaez2011contour} after removing
image~\textrm{302003.jpg}, which overlaps with one of the test images.
$16 \times 16$ versions of the patches are build using the Matlab function \textrm{imresize},
and upscaled back to $32 \times 32$ by using bicubic interpolation; then, the
goal is to predict high-resolution images from blurry bicubic
interpolations.

The blurry estimates are processed by a $9$-layer network, with $3\times 3$ patches
and~$128$ filters at every layer without linear pooling and zero-padding. 
Pixel values are predicted with a linear model applied to the $128$-dimensional 
vectors present at every pixel location of the last layer, and we use the square loss to measure the fit.
The optimization procedure and the kernels~$\kappa_j$ are identical to the ones used for
processing the SVHN dataset in the classification task.
The pipeline also includes a pre-processing step, where we remove from input images a local mean component obtained
by convolving the images with a $5 \times 5$ averaging box filter;
the mean component is added back after up-scaling.

For the evaluation, we consider three datasets: Set5 and Set14 are standard
for super-resolution; Kodim is the Kodak Image database, available at
\url{http://r0k.us/graphics/kodak/}, which contains high-quality images with no
compression or demoisaicing artefacts.  The evaluation procedure follows
~\cite{dong2014learning,dong2015image,timofte2013anchored,wang2015deep}
by using the code from the author's web page. We present quantitative results
in Table~\ref{table:psnr}.  For x3 upscaling, we simply used twice our model
learned for x2 upscaling, followed by a 3/4 downsampling. This is clearly
suboptimal since our model is not trained to up-scale by a factor~3, but this 
naive approach still outperforms other
baselines~\cite{dong2014learning,dong2015image,wang2015deep} that are trained
end-to-end. Note that~\cite{wang2015deep} also proposes a data augmentation
scheme at test time that slightly improves their results.
In Appendix~\ref{appendix:superres}, we also present a visual comparison
between our approach and~\cite{dong2015image}, whose pipeline is the closest to
ours, up to the use of a supervised kernel machine instead of CNNs.

\begin{table}[hbtp]
   \caption{Reconstruction accuracy for super-resolution in PSNR (the higher, the better). All CNN approaches are without data augmentation at test time. See Appendix~\ref{appendix:superres} for the SSIM quality measure.}\label{table:psnr}
   \vspace*{-0.1cm}
   \centering
   \renewcommand\tabcolsep{0.15cm}
   \footnotesize
   \begin{tabular}{| c|c||c|c|c|c|c|c|c||c|}
      \hline
      Fact. & Dataset & Bicubic & SC~\cite{zeyde2010single} & ANR~\cite{timofte2013anchored} & A+\cite{timofte2013anchored} & CNN1~\cite{dong2014learning} & CNN2~\cite{dong2015image} & CSCN~\cite{wang2015deep}  & SCKN \\
      \hline
      \multirow{3}{*}{x2} & Set5  & 33.66 & 35.78 & 35.83 & 36.54 & 36.34 & 36.66 & 36.93 & \textbf{37.07} \\
                          & Set14 & 30.23 & 31.80 & 31.79 & 32.28 & 32.18 & 32.45 & 32.56 & \textbf{32.76} \\
                          & Kodim & 30.84 & 32.19 & 32.23 & 32.71 & 32.62 & 32.80 & 32.94 & \textbf{33.21} \\
      \hline
      \multirow{3}{*}{x3} & Set5  & 30.39 & 31.90 & 31.92 & 32.58 & 32.39 & 32.75 & \textbf{33.10} & 33.08 \\
                          & Set14 & 27.54 & 28.67 & 28.65 & 29.13 & 29.00 & 29.29 & 29.41 & \textbf{29.50} \\
                          & Kodim & 28.43 & 29.21 & 29.21 & 29.57 & 29.42 & 29.64 & 29.76 & \textbf{29.88} \\
      \hline
   \end{tabular}
   \vspace*{-0.2cm}
\end{table}

% \section{Conclusion}
% \input{ccl.tex}

\subsubsection*{Acknowledgments}
This work was supported by ANR (MACARON project ANR-14-CE23-0003-01).

%\section*{References}

\newpage
{
\small
%\addtolength{\bibsep}{-2.7pt}
\bibliographystyle{plain}
\bibliography{abbrev,main}
}

\appendix
\section{Orthogonal Projection on the Finite-Dimensional Subspace $\Fcal_1$}\label{appendix:A}
First, we remark that the kernel $K_1$ is homogeneous such that for every patch~$\x$ and scalar $\gamma > 0$,
\begin{displaymath}
   \varphi_1(\gamma \x) = \gamma \varphi_1(\x).
\end{displaymath}
Thus, we may assume $\x$ to have unit $\ell_2$-norm without loss of generality
and perform the projection on~$\Fcal_1 =
\text{Span}(\varphi(\z_1),\ldots,\varphi(\z_{p_1}))$ of the normalized patch,
before applying the inverse rescaling.

Then, let us denote by~$f_{\x}$ the orthogonal projection of a patch~$\x$ with unit $\ell_2$-norm defined as
\begin{equation*}
    f_{\x} \defin \argmin_{f \in \Fcal_1} \| \varphi_1(\x) - f \|_{\Hcal_1}^2,
\end{equation*}
which is equivalent to 
\begin{equation*}
   f_{\x} \defin \sum_{j=1}^{p_1} \alpha_j^\star \varphi_1(\z_j) ~~~\text{with}~~~ \alphab^\star \in \argmin_{\alphab \in \Real^{p_1}}\left\| \varphi_1(\x) - \sum_{j=1}^{p_1} \alpha_j \varphi_1(\z_j)\right\|_{\Hcal_1}^2.
\end{equation*}
After short calculation, we obtain
\begin{equation*}
   f_{\x} = \sum_{j=1}^{p_1} \alpha_j^\star \varphi_1(\z_j) ~~~\text{with}~~~ \alphab^\star \in \argmin_{\alphab \in \Real^{p_1}} \left [ 1 - 2 \alphab^\top \kappa_1(\z^\top \x) + \alphab^\top \kappa_1(\Z^\top\Z) \alphab\right],
\end{equation*}
since the vectors~$\z_j$ provided by the spherical K-means algorithm have unit
$\ell_2$-norm. Assuming $\kappa_1(\Z^\top \Z)$ to be invertible, we have $\alphab^\star =
\kappa_1(\Z^\top\Z)^{-1} \kappa_1(\Z^\top \x)$.  
After projection, normalized patches $\x$, $\x'$ may be parametrized by
$\alphab^\star = \kappa_1(\Z^\top\Z)^{-1} \kappa_1(\Z^\top \x)$ and
$\alphab^{\prime\star} = \kappa_1(\Z^\top\Z)^{-1} \kappa_1(\Z^\top \x')$, respectively. Then, we have 
\begin{displaymath}
\langle f_{\x}, f_{\x'} \rangle_{\Hcal_1} = \alphab^{\star\top} \kappa_1(\Z^\top \Z) \alphab^{\prime \star} =  \left\langle \psi_1(\x), \psi_1(\x') \right\rangle, 
\end{displaymath}
which is the desired result.

When $\kappa_1(\Z^\top\Z)$ is not invertible or simply badly conditioned, it is also common to use instead
\begin{displaymath}
   \psi_1(\x) =  \left(\kappa_1(\Z^\top\Z)+ \varepsilon \I \right)^{-1/2} \kappa_1(\Z^\top \x),
\end{displaymath}
where $\varepsilon > 0$ is a small regularization that improves the condition
number of $\kappa_1(\Z^\top\Z)$. Such a modification can be interpreted as performing a
slightly regularized projection onto the finite-dimensional subspace~$\Fcal_1$.

\section{Computation of the Gradient with Respect to the Filters}\label{appendix:proof}
To compute the gradient of the loss function, we use Lemma~\ref{lemma:backprop}
and start by analyzing the effect of perturbing every quantity involved
in~(\ref{eq:forward}) such that we may obtain the desired relations~(\ref{eq:perturb})
and~(\ref{eq:perturb2}). Before proceeding, we recall the definition of the set $\Zcal+\Ecal=\{
\Z_1+\varepsilonb_1, \ldots, \Z_k + \varepsilonb_k \}$ and the precise definition of the 
Landau notation $o(\|\Ecal\|)$, which we use in~(\ref{eq:perturb}). Here, it simply means
a quantity that is negligible in front of the norm $\|\Ecal\|= \sum_{j=1}^k
\|\varepsilonb_j\|_{\text{F}}$---that is, 
\begin{displaymath}
   \lim_{\Ecal \to 0} \frac{\left\|I_{j}^{\Zcal+\Ecal} - I_{j}^{\Zcal} - \Delta I_{j}^{\Zcal,\Ecal}\right\|_{\text{F}}}{\|\Ecal\|} = 0.
\end{displaymath}
Then, we start by initializing a recursion: $I_0^\Zcal$ is unaffected by the
perturbation and thus $\Delta I_{0}^{\Zcal,\Ecal}=0$. Consider now an index~$j
> 0$ and assume that~(\ref{eq:perturb}) holds for $j-1$ with $\Delta I_{\jmone}^{\Zcal,\Ecal} = O(\|\Ecal\|)$.

First, we remark that 
\begin{displaymath}
   \E_j(I_{\jmone}^{\Zcal+\Ecal}) = \E_j(I_{\jmone}^{\Zcal}) + \E_j(\Delta I_{\jmone}^{\Zcal,\Ecal}) + o(\|\Ecal\|).
\end{displaymath}
Then, the diagonal matrix~$\S_j$ becomes after perturbation
\begin{displaymath}
   \S_j + \underbrace{\S_j^{-1} \odot \left( \E_j(I_{\jmone}^{\Zcal})^\top   \E_j(\Delta I_{\jmone}^{\Zcal,\Ecal}) \right)}_{\Delta \S_j} + o(\|\Ecal\|).
\end{displaymath}
The inverse diagonal matrix~$\S_j^{-1}$ becomes
\begin{displaymath}
   \S_j^{-1} \underbrace{- \S_j^{-3} \odot \left( \E_j(I_{\jmone}^{\Zcal})^\top   \E_j(\Delta I_{\jmone}^{\Zcal,\Ecal}) \right)}_{\Delta \S_j^{-1}} + o(\|\Ecal\|),
\end{displaymath}
and the matrix $\A_j$ becomes
\begin{displaymath}
   \begin{split}
      \!\kappa_j\left((\Z_j \!+ \!\varepsilonb_j)^\top (\Z_j \!+ \!\varepsilonb_j)\right)^{-\frac{1}{2}} \! &= \kappa_j\left(\Z_j^\top\Z_j \!+ \!\varepsilonb_j^\top\Z_j \!+ \!\Z_j^\top \varepsilonb_j + o(\|\varepsilonb_j\|_{\text{F}}))\right)^{-\frac{1}{2}} \\
      & = 
       \left( \kappa_j\left( \Z_j^\top \Z_j\right) \!+ \!\kappa_j'\left( \Z_j^\top \Z_j\right) \odot \left(\Z_j^\top \varepsilonb_j \!+\! \varepsilonb_j^\top \Z_j\right) \!+\! o(\|\Ecal\|) \right)^{-\frac{1}{2}}  \\
       & =  \A_j^{\frac{1}{2}}\left( \I \!+\! \A_j \left(\kappa_j'\left( \Z_j^\top \Z_j\right) \odot \left(\Z_j^\top \varepsilonb_j \!+\! \varepsilonb_j^\top \Z_j\right)\right)\A_j \!+\! o(\|\Ecal\|) \right)^{-{1}/{2}}\A_j^{\frac{1}{2}} \\
       & =  \A_j \underbrace{- \frac{1}{2}\A_j^{\frac{3}{2}}\left( \kappa_j'\left( \Z_j^\top \Z_j\right) \odot \left(\Z_j^\top \varepsilonb_j \!+\! \varepsilonb_j^\top \Z_j\right)\right)\A_j^{\frac{3}{2}}}_{\Delta \A_j}  + o(\|\Ecal\|),
   \end{split}
\end{displaymath}
where we have used the relation $(\I+\Q)^{-1/2}=\I-\frac{1}{2}\Q + o(\|\Q\|_{\text{F}})$.
Note that the quantities $\Delta \A_j, \Delta \S_j, \Delta \S_j^{-1}$ that we have introduced are all $O(\|\Ecal\|)$.
Then, by replacing the quantities $\A_j, \S_j, \S_j^{-1}, \I_{\jmone}$ by their
perturbed versions in the definition of~$I_j$ given in~(\ref{eq:forward}), we
obtain that $I_j^{\Zcal+\Ecal}$ is equal to
\begin{displaymath}
(\A_j \!+\! \Delta \A_j) \kappa_j\!\left( (\Z_j \!+\! \varepsilonb_j)^\top\left(\E_j(I_{\jmone}^{\Zcal})\!+\!  \E_j(\Delta I_{\jmone}^{\Zcal,\Ecal})\right) (\S_j^{-1} \!+\! \Delta \S_j^{-1}) \right)(\S_j \!+\! \Delta \S_j)\P_j \!+\! o(\|\Ecal\|).
\end{displaymath}
Then, after short calculation, we obtain the desired relation $I_j^{\Zcal+\Ecal} = I_j^{\Zcal+\Ecal} +  \Delta I_{j}^{\Zcal,\Ecal} + o(\|\Ecal\|)$ with 
\begin{displaymath}
   \begin{split}
      \Delta I_{j}^{\Zcal,\Ecal}  = &  \Delta \A_j \kappa_j (\Z_j^\top \E_j(I_{\jmone}^{\Zcal}) \S_j^{-1})\S_j\P_j \\
                              & + \A_j \left(\kappa_j'(\Z_j^\top \E_j(I_{\jmone}^{\Zcal}) \S_j^{-1}) \odot (\varepsilonb_j^\top \E_j(I_{\jmone}^{\Zcal}))\right)\P_j \\
                              & + \A_j \left(\kappa_j'(\Z_j^\top \E_j(I_{\jmone}^{\Zcal}) \S_j^{-1}) \odot (\Z_j^\top \E_j(\Delta I_{\jmone}^{\Zcal,\Ecal}))\right)\P_j \\
                              & + \A_j \left(\kappa_j'(\Z_j^\top \E_j(I_{\jmone}^{\Zcal}) \S_j^{-1}) \odot (\Z_j^\top \E_j(I_{\jmone}^{\Zcal}) \Delta\S_j^{-1}\S_j)\right)\P_j \\
                              & + \A_j \kappa_j (\Z_j^\top \E_j(I_{\jmone}^{\Zcal}) \S_j^{-1})\Delta\S_j\P_j.
   \end{split}
\end{displaymath}
First, we remark that $\Delta I_{j}^{\Zcal,\Ecal} = O(\|\Ecal\|)$, which is one of the induction hypothesis we need. Then,
after plugging in the values of $\Delta \A_j, \Delta \S_j, \Delta \S_j^{-1}$, and with further simplification, we obtain
\begin{multline*}
    \Delta I_{j}^{\Zcal,\Ecal} = - \frac{1}{2}\A_j^{\frac{3}{2}}\left( \kappa_j'\left( \Z_j^\top \Z_j\right) \odot \left(\Z_j^\top \varepsilonb_j + \varepsilonb_j^\top \Z_j\right)\right)\A_j^{\frac{1}{2}}I_j^{\Zcal} \\ 
    + \A_j\left(\kappa'_j(\Z_j^\top \E_j(I_{\jmone}^\Zcal) \S_j^{-1})\odot \left(\varepsilonb_j^\top \E_j(I_{\jmone}^{\Zcal})\right)\right)\P_j \\
  + \A_j\left(\kappa'_j(\Z_j^\top \E_j(I_{\jmone}^\Zcal) \S_j^{-1})\odot \left(  \Z_j^\top \E_j(\Delta I_{\jmone}^{\Zcal,\Ecal}) \right) \right)\P_j \\ 
  - \A_j\left(\kappa'_j(\Z_j^\top \E_j(I_{\jmone}^\Zcal) \S_j^{-1})\odot \left(\Z_j^\top \E_j(I_{\jmone}^{\Zcal})  \left( \S_j^{-2} \odot \left( \E_j(I_{\jmone}^{\Zcal})^\top   \E_j(\Delta I_{\jmone}^{\Zcal,\Ecal}) \right)\right)\right) \right)\P_j \\ 
    +  M_j^\Zcal \left( \S_j^{-2} \odot \left( \E_j(I_{\jmone}^{\Zcal})^\top   \E_j(\Delta I_{\jmone}^{\Zcal,\Ecal}) \right)\right)\P_j,\\
\end{multline*}
where $M_j^\Zcal$ is the $j$-th feature map of~$I_0$ before the $j$-th linear pooling step---that is, $I_j^\Zcal = M_j^\Zcal \P_j$.
We now see that $\Delta I_{j}^{\Zcal,\Ecal}$ is linear in~$\varepsilonb_j$ and~$\Delta I_{\jmone}^{\Zcal,\Ecal}$, which guarantees that there exist two linear
functions~$g_j, h_j$ that satisfy~(\ref{eq:perturb2}).
More precisely, we want for all matrix~$\U$ of the same size as~$\Delta I_{j}^{\Zcal,\Ecal}$
\begin{multline*}
   \langle \varepsilonb_j, g_j(\U) \rangle = \left\langle   -
   \frac{1}{2}\A_j^{\frac{3}{2}}\left( \kappa_j'\left( \Z_j^\top \Z_j\right)
   \odot \left(\Z_j^\top \varepsilonb_j + \varepsilonb_j^\top
\Z_j\right)\right)\A_j^{\frac{1}{2}}I_j^{\Zcal}, \U \right\rangle \\
+ \left\langle \A_j\left(\kappa'_j(\Z_j^\top \E_j(I_{\jmone}^\Zcal) \S_j^{-1})\odot \left(\varepsilonb_j^\top \E_j(I_{\jmone}^{\Zcal})\right)\right)\P_j, \U \right\rangle,
\end{multline*}
and
\begin{multline*}
   \langle \Delta I_{\jmone}^{\Zcal,\Ecal}, h_j(\U) \rangle = \left\langle \A_j\left(\kappa'_j(\Z_j^\top \E_j(I_{\jmone}^\Zcal) \S_j^{-1})\odot \left(  \Z_j^\top \E_j(\Delta I_{\jmone}^{\Zcal,\Ecal}) \right) \right)\P_j , \U \right\rangle \\
  - \left\langle \A_j\left(\kappa'_j(\Z_j^\top \E_j(I_{\jmone}^\Zcal) \S_j^{-1})\odot \left(\Z_j^\top \E_j(I_{\jmone}^{\Zcal})  \left( \S_j^{-2} \odot \left( \E_j(I_{\jmone}^{\Zcal})^\top   \E_j(\Delta I_{\jmone}^{\Zcal,\Ecal}) \right)\right)\right) \right)\P_j, \U \right\rangle \\ 
    + \left\langle  M_j^\Zcal \left( \S_j^{-2} \odot \left( \E_j(I_{\jmone}^{\Zcal})^\top   \E_j(\Delta I_{\jmone}^{\Zcal,\Ecal}) \right)\right)\P_j,\U \right\rangle.
\end{multline*}
Then, it is easy to obtain the form of~$g_j, h_j$ given in~(\ref{eq:gjhj}), by using in the right order the following elementary calculus rules: (i) $\langle \U\V, \W \rangle = \langle \U, \W\V^\top \rangle = \langle \V, \U^\top\W \rangle$, (ii) $\langle \U, \V \rangle = \langle \U^\top, \V^\top \rangle$, (iii) $\langle \U \odot \V, \W\rangle = \langle \U , \V \odot \W \rangle$ for any matrices $\U,\V,\W$ of appropriate sizes, and also (iv) $\langle \E_j(\U), \V\rangle=\langle \U, \E_j^\star(\V) \rangle$, by definition of the adjoint operator. We conclude by induction.

\section{Preconditioning Heuristic on the Sphere}\label{appendix:precond}
In this section, we present a preconditioning heuristic for optimizing 
over the sphere ${\mathbb S}^{p-1}$, inspired by second-order (Newton)
optimization techniques on smooth manifolds~\cite{absil2009optimization}.
Following~\cite{absil2009optimization}, we will consider gradient descent steps 
on the manifold. A fundamental operation is thus
the projection operator $P_\z$ onto the tangent space at a
point~$\z$. This operator is defined for the sphere by
\begin{displaymath}
   P_\z[\u] = (\I-\z\z^\top)\u,
\end{displaymath}
for any vector~$\u$ in~$\Real^p$. Another important operator is the Euclidean projection on
${\mathbb S}^{p-1}$, which was denoted by~$\proj$ in
previous parts of the paper. 
\paragraph{Gradient descent on the sphere~${\mathbb S}^{p-1}$ is equivalent to the projected gradient descent in~$\Real^p$.}~\\
When optimizing on a manifold, the natural descent direction is the projected gradient $P_\z \nabla L(\z)$. In the case of the sphere, a gradient
step on the manifold is equivalent to a classical projected gradient descent step in~$\Real^p$ with particular step size:
\begin{displaymath}
   \begin{split}
      \proj[\z - \eta P_\z[\nabla L(\z)]] & = \proj\left[\z - \eta \left(\I-\z\z^\top\right)\nabla L(\z)\right] \\
                                          & = \proj\left[\left(1 + \eta\z^\top\nabla L(\z)\right)\z - \eta\nabla L(\z)\right] \\
                                          & = \proj\left[\z - \frac{\eta}{1 + \eta\z^\top\nabla L(\z)}\nabla L(\z)\right]. 
   \end{split}
\end{displaymath}
\paragraph{In~$\Real^p$ with no constraint, pre-conditioning is equivalent to performing a change of variable.}~\\
For \emph{unconstrained} optimization in~$\Real^p$, faster convergence is usually achieved when one
has access to an estimate of the inverse of the Hessian~$\nabla^2 L(\z)$---assuming twice differentiability---and using the
descent direction $(\nabla^2 L(\z))^{-1}\nabla L(\z)$ instead of~$\nabla L(\z)$; then, we
obtain a Newton method. When the exact Hessian is not available, or too costly
to compute and/or invert, it is however common to use instead a constant estimate of
the inverse Hessian, denoted here by~$\Q$, which we call pre-conditioning
matrix. Finding an appropriate matrix~$\Q$ is difficult in general, but for learning linear
models, a typical choice is to use the inverse covariance matrix of the
data (or one approximation). In that case, the preconditioned gradient descent
step consists of the update $\z - \eta \Q\nabla L(\z)$. Such a matrix~$\Q$
is defined similarly in the context of convolutional kernel networks, as
explained in the main part of the paper. A useful interpretation of
preconditioning is to see it as optimizing after a change of variable. Define
indeed the objective 
$$\tilde{L}(\w) = L(\Q^{1/2} \w).$$
Then, minimizing $\tilde{L}$ is equivalent to minimizing~$L$ with respect
to~$\z$, with the relation $\z=\Q^{1/2}\w$. Moreover, when there is no
constraint on~$\z$ and~$\w$, the regular gradient descent algorithm on~$\tilde{L}$ is
equivalent to the preconditioned gradient descent on~$L$:
\begin{displaymath}
   \begin{split}
      \w \leftarrow \w - \eta \nabla \tilde{L}(\w) & \Longleftrightarrow  \w \leftarrow \w - \eta \Q^{1/2}\nabla L(\Q^{1/2}\w) \\
                                                   & \Longleftrightarrow  \z \leftarrow \z - \eta \Q\nabla L(\z)~~~\text{with}~~~\z=\Q^{1/2}\w.
\end{split}
\end{displaymath}
We remark that the Hessian~$\nabla^2 \tilde{L}(\w)$ is equal to
$\Q^{1/2}\nabla^2{L}(\Q^{1/2}\w) \Q^{1/2}$, which is equal to identity when
$\Q$ coincides with the inverse Hessian of~$L$. In general, this is of course
not the case, but the hope is to obtain a Hessian $\nabla^2 \tilde{L}$ that is better
conditioned than $\nabla^2{L}$, thus resulting in faster convergence.

\paragraph{Preconditioning on a smooth manifold requires some care.}~\\
Unfortunately, using second-order information (or simply a pre-conditioning
matrix) when optimizing over a constraint set or over a smooth manifold is not as
simple as optimizing in~$\Real^p$ since the quantities $\Q\nabla L(\z), P_z[\Q\nabla L(\z)]
,\Q P_z[\nabla L(\z)]$ may not be feasible descent directions.
However, the point of view that sees pre-conditioning as
a change of variable will give us the right direction to follow.

Optimizing~$L$ on~$\Sbb^{p-1}$ is in fact equivalent to optimizing~$\tilde{L}$ on the
smooth manifold
\begin{displaymath}
   \tilde{\Sbb}^{p-1} = \left\{  \w \in \Real^p : \|\Q^{1/2}\w\|_2 = 1 \right\},
\end{displaymath}
which represents an ellipsoid. The tangent plane at a point~$\w$ of $\tilde{\Sbb}^{p-1}$ being defined by
the normal vector $\Q\w/\|\Q\w\|_2$, it is then possible to introduce the projection operator~$\tilde{P}_\w$ on the
tangent space:
\begin{displaymath}
   \tilde{P}_\w[\u] = \left(\I - \frac{\Q \w \w^\top \Q}{\w^\top \Q^2 \w}\right) \u.
\end{displaymath}
Then, we may define the gradient descent step rule on $\tilde{\Sbb}^{p-1}$ as
\begin{displaymath}
      \w \leftarrow \projb\left[\w - \eta \tilde{P}_\w\left[\nabla \tilde{L}(\w)\right]\right] = \projb\left[\w - \eta \left(\I - \frac{\Q \w \w^\top \Q}{\w^\top \Q^2 \w}\right) \Q^{1/2}\nabla L(\Q^{1/2}\w)\right]. 
\end{displaymath}
With the change of variable $\z=\Q^{1/2}\w$, this is equivalent to
\begin{displaymath}
   \z \leftarrow \proj\left[\z - \eta \left(\I - \frac{\Q \z \z^\top}{\z^\top \Q \z}\right) \Q\nabla L(\z)\right].
\end{displaymath}
This is exactly the update rule we have chosen in our paper, as a heuristic in a stochastic setting.

\section{Additional Results for Image Super-Resolution}\label{appendix:superres}
We present a quantitative comparison in Table~\ref{table:ssim} using
the structural similarity index measure (SSIM), which is known to better
reflect the quality perceived by humans than the PSNR; it is commonly used 
to evaluate the quality of super-resolution methods, see~\cite{dong2015image,timofte2013anchored,wang2015deep}. Then, we present a visual
comparison between several approaches in Figures~\ref{fig:visual},~\ref{fig:visual2}, and~\ref{fig:visual3}. We focus notably on the classical convolutional neural network
of~\cite{dong2015image} since our pipeline essentially differs in the use of our supervised kernel machine instead of convolutional neural networks.
After subjective evaluation, we observe that both methods perform
equally well in textured areas. However, our approach recovers better
thin high-frequency details, such as the eyelash of the baby in the first image.
By zooming on various parts, it is easy to notice similar differences in other images.
We also observed a few ghosting artefacts near object boundaries with the method of~\cite{dong2015image},
which is not the case with our approach.
\begin{table}[hbtp]
   \caption{Reconstruction accuracy of various super-resolution approaches. The numbers represent the structural similarity index (SSIM), the higher, the better.}\label{table:ssim}
   \centering
   \renewcommand\tabcolsep{0.15cm}
   \footnotesize
   \begin{tabular}{| c|c||c|c|c|c|c|c|c||c|}
      \hline
      Fact. & Dataset & Bicubic & SC~\cite{zeyde2010single} & ANR~\cite{timofte2013anchored} & A+\cite{timofte2013anchored} & CNN1~\cite{dong2014learning} & CNN2~\cite{dong2015image} & CSCN~\cite{wang2015deep}  & SCKN \\
      \hline
      \multirow{3}{*}{x2} & Set5  & 0.9299 & 0.9492 & 0.9499 & 0.9544 & 0.9521 & 0.9542 & 0.9552 & \textbf{0.9580} \\
                          & Set14 & 0.8689 & 0.8989 & 0.9004 & 0.9056 & 0.9037 & 0.9067 & 0.9074 & \textbf{0.9115} \\
                          & Kodim & 0.8684 & 0.8990 & 0.9007 & 0.9075 & 0.9043 & 0.9068 & 0.9104 & \textbf{0.9146} \\
      \hline
      \multirow{3}{*}{x3} & Set5  & 0.8677 & 0.8959 & 0.8959 & 0.9088 & 0.9025 & 0.9090 & 0.9144 & \textbf{0.9165} \\
                          & Set14 & 0.7741 & 0.8074 & 0.8092 & 0.8188 & 0.8148 & 0.8215 & 0.8238 & \textbf{0.8297} \\
                          & Kodim & 0.7768 & 0.8066 & 0.8084 & 0.8175 & 0.8109 & 0.8174 & 0.8222 & \textbf{0.8283} \\
      \hline
   \end{tabular}
\end{table}

\begin{figure}
   \renewcommand\tabcolsep{0.1cm}
   \begin{tabular}{cccc}
      \includegraphics[width=0.24\linewidth]{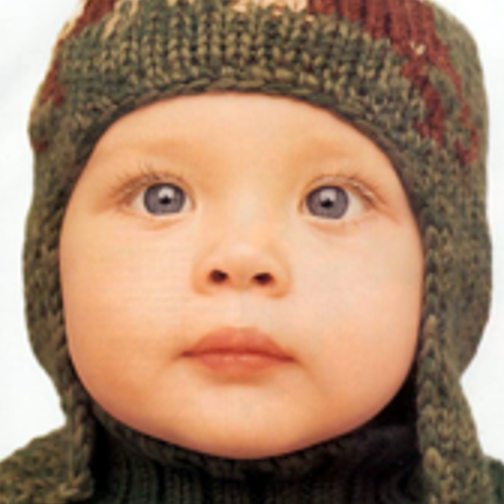} &
      \includegraphics[width=0.24\linewidth]{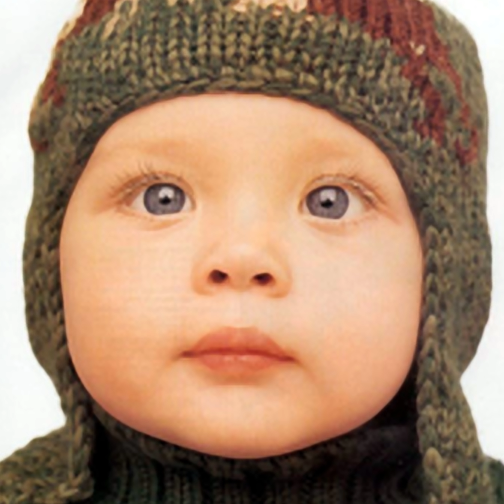} &
      \includegraphics[width=0.24\linewidth]{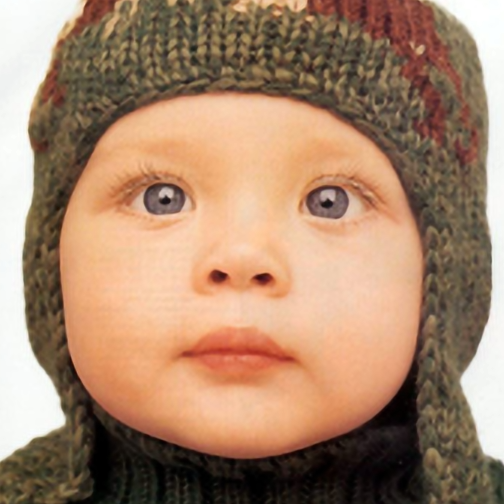} &
      \includegraphics[width=0.24\linewidth]{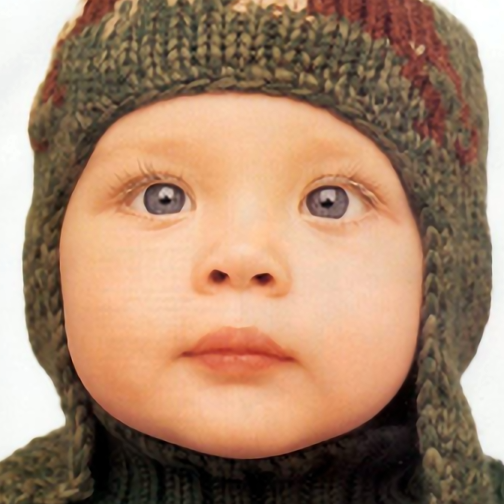} \\
      \includegraphics[width=0.24\linewidth,trim=115 280 300 155,clip]{figures/baby_GT.bmp/fact3_bicubic_baby_GT.png} &
      \includegraphics[width=0.24\linewidth,trim=115 280 300 155,clip]{figures/baby_GT.bmp/sc.png} &
      \includegraphics[width=0.24\linewidth,trim=115 280 300 155,clip]{figures/baby_GT.bmp/fact3_srcnn_baby_GT.png} &
      \includegraphics[width=0.24\linewidth,trim=115 280 300 155,clip]{figures/baby_GT.bmp/fact3_exp_baby_GT.png} \\
      \includegraphics[width=0.24\linewidth]{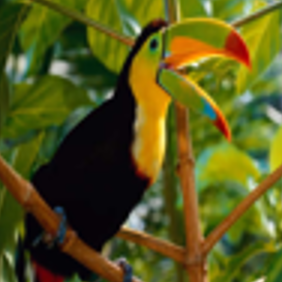} &
      \includegraphics[width=0.24\linewidth]{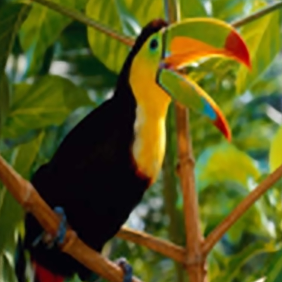} &
      \includegraphics[width=0.24\linewidth]{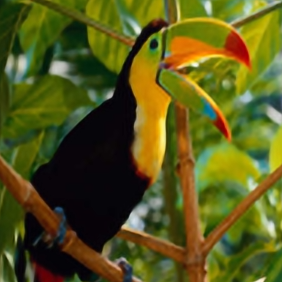} &
      \includegraphics[width=0.24\linewidth]{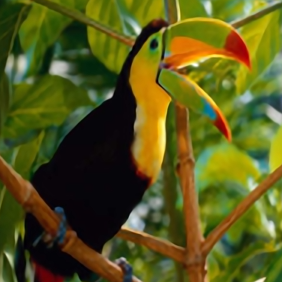} \\
      \includegraphics[width=0.24\linewidth,trim=120 170 70 35,clip]{figures/bird_GT.bmp/fact3_bicubic_bird_GT.png} &
      \includegraphics[width=0.24\linewidth,trim=120 170 70 35,clip]{figures/bird_GT.bmp/sc.png} &
      \includegraphics[width=0.24\linewidth,trim=120 170 70 35,clip]{figures/bird_GT.bmp/fact3_srcnn_bird_GT.png} &
      \includegraphics[width=0.24\linewidth,trim=120 170 70 35,clip]{figures/bird_GT.bmp/fact3_exp_bird_GT.png} \\
      \includegraphics[width=0.24\linewidth]{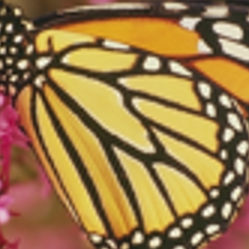} &
      \includegraphics[width=0.24\linewidth]{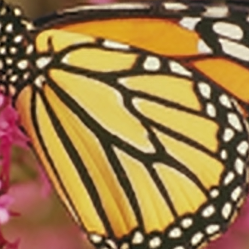} &
      \includegraphics[width=0.24\linewidth]{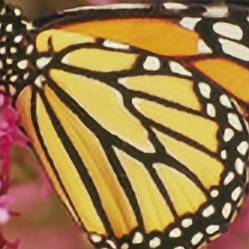} &
      \includegraphics[width=0.24\linewidth]{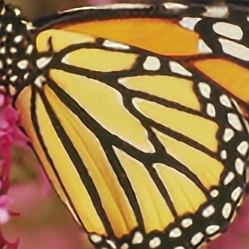} \\
      \includegraphics[width=0.24\linewidth,trim=40 20 100 120,clip]{figures/butterfly_GT.bmp/fact3_bicubic_butterfly_GT.png} &
      \includegraphics[width=0.24\linewidth,trim=40 20 100 120,clip]{figures/butterfly_GT.bmp/sc.png} &
      \includegraphics[width=0.24\linewidth,trim=40 20 100 120,clip]{figures/butterfly_GT.bmp/fact3_srcnn_butterfly_GT.png} &
      \includegraphics[width=0.24\linewidth,trim=40 20 100 120,clip]{figures/butterfly_GT.bmp/fact3_exp_butterfly_GT.png} \\
      Bicubic & Sparse coding~\cite{zeyde2010single} & CNN2~\cite{dong2015image} & SCKN (Ours)
   \end{tabular}
   \caption{Visual comparison for x3 image up-scaling. 
      Each column corresponds to a different method (see bottom row). RGB images are converted to the YCbCr color space and the
      up-scaling method is applied to the luminance channel only. Color
   channels are up-scaled using bicubic interpolation for visualization purposes. CNN2 and SCKN perform similarly in textured areas, but SCKN provides significantly sharper artefact-free edges (see in particular the butterfly image). Best seen by zooming on a computer screen with an appropriate PDF viewer that does not smooth the image content.}\label{fig:visual}
\end{figure} 

\begin{figure}
   \renewcommand\tabcolsep{0.1cm}
   \begin{tabular}{cccc}
      \includegraphics[width=0.24\linewidth]{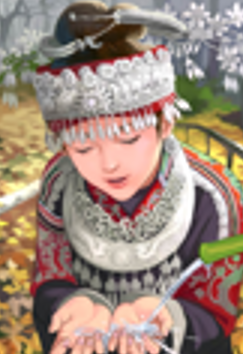} &
      \includegraphics[width=0.24\linewidth]{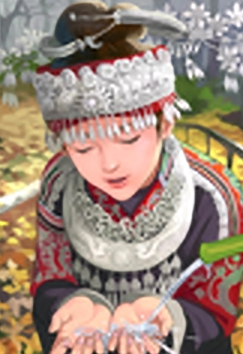} &
      \includegraphics[width=0.24\linewidth]{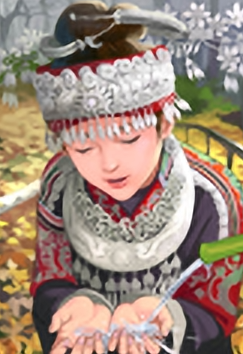} &
      \includegraphics[width=0.24\linewidth]{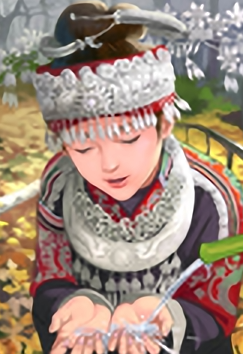} \\
      \includegraphics[width=0.24\linewidth,trim=160 140 0 130,clip]{figures/comic.bmp/fact3_bicubic_comic.png} &
      \includegraphics[width=0.24\linewidth,trim=160 140 0 130,clip]{figures/comic.bmp/sc.png} &
      \includegraphics[width=0.24\linewidth,trim=160 140 0 130,clip]{figures/comic.bmp/fact3_srcnn_comic.png} &
      \includegraphics[width=0.24\linewidth,trim=160 140 0 130,clip]{figures/comic.bmp/fact3_exp_comic.png} \\
      \includegraphics[width=0.24\linewidth]{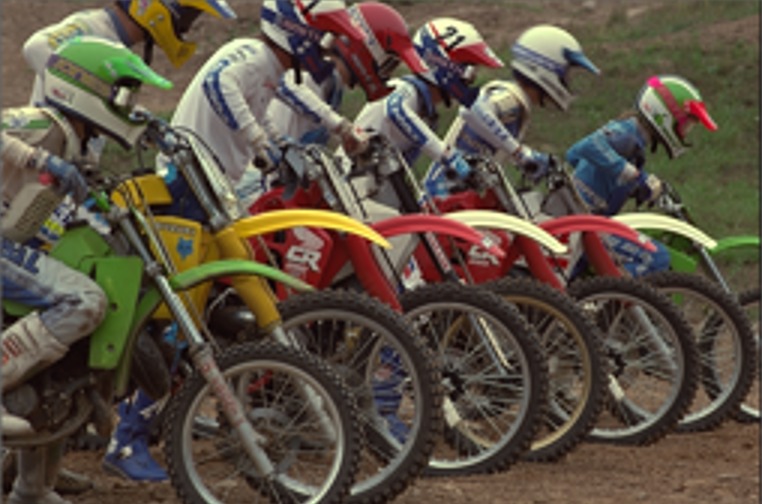} &
      \includegraphics[width=0.24\linewidth]{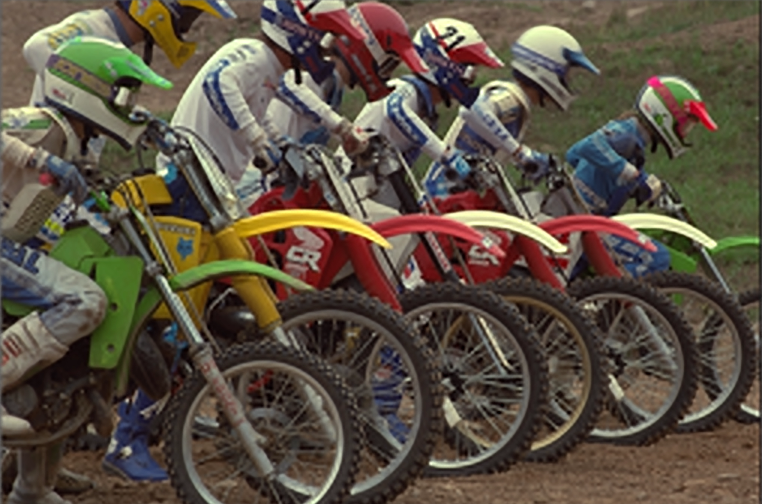} &
      \includegraphics[width=0.24\linewidth]{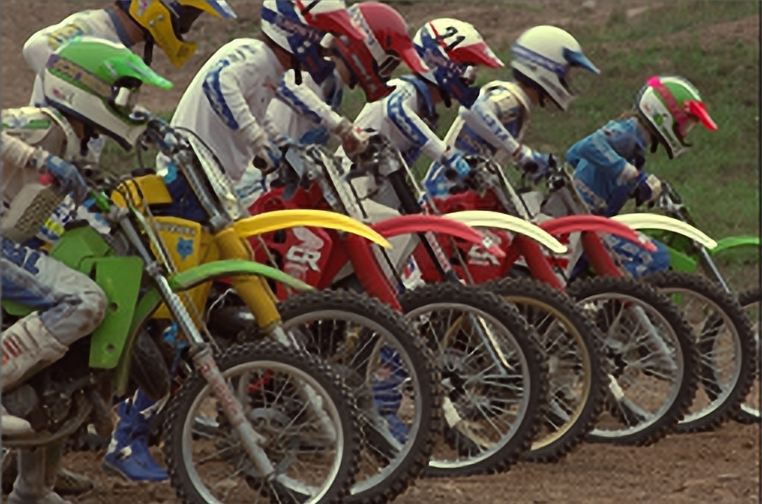} &
      \includegraphics[width=0.24\linewidth]{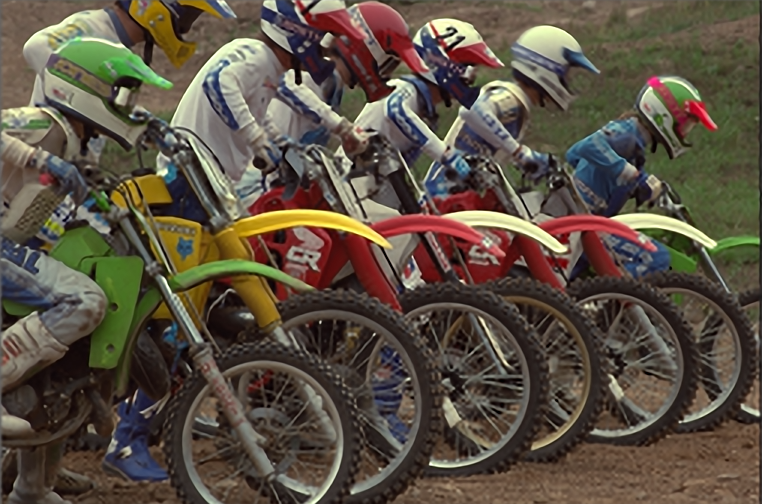} \\
      \includegraphics[width=0.24\linewidth,trim=320 200 280 200,clip]{figures/kodim05.png/fact3_bicubic_kodim05.jpg} &
      \includegraphics[width=0.24\linewidth,trim=320 200 280 200,clip]{figures/kodim05.png/sc.png} &
      \includegraphics[width=0.24\linewidth,trim=320 200 280 200,clip]{figures/kodim05.png/fact3_srcnn_kodim05.png} &
      \includegraphics[width=0.24\linewidth,trim=320 200 280 200,clip]{figures/kodim05.png/fact3_exp_kodim05.png} \\
      \includegraphics[width=0.24\linewidth]{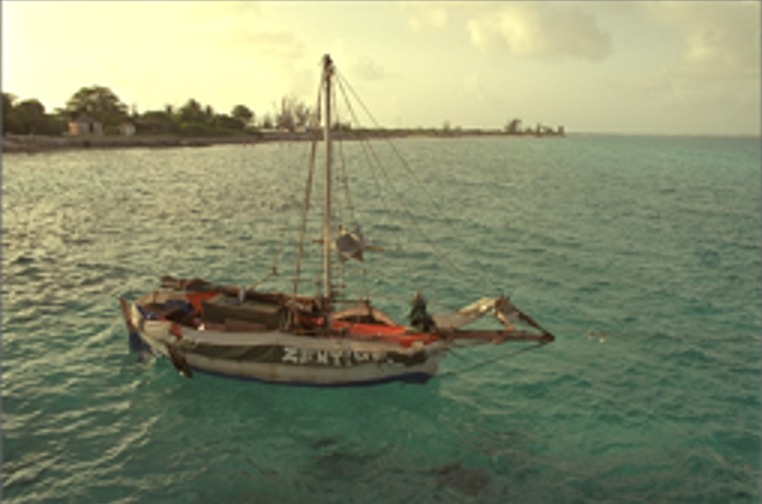} &
      \includegraphics[width=0.24\linewidth]{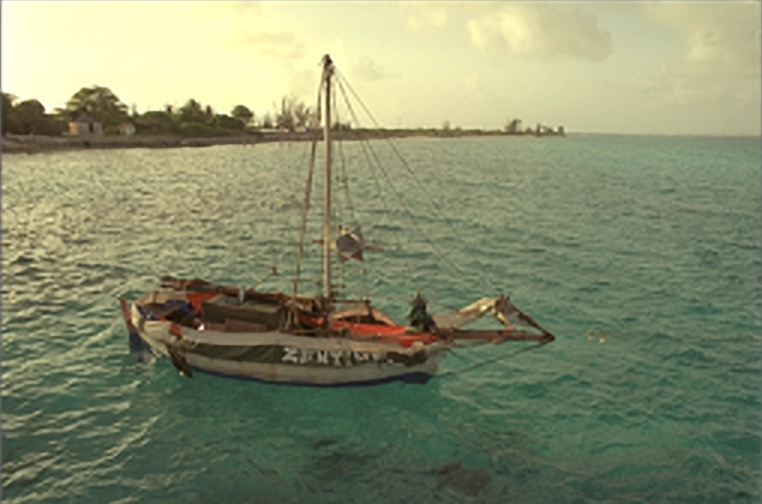} &
      \includegraphics[width=0.24\linewidth]{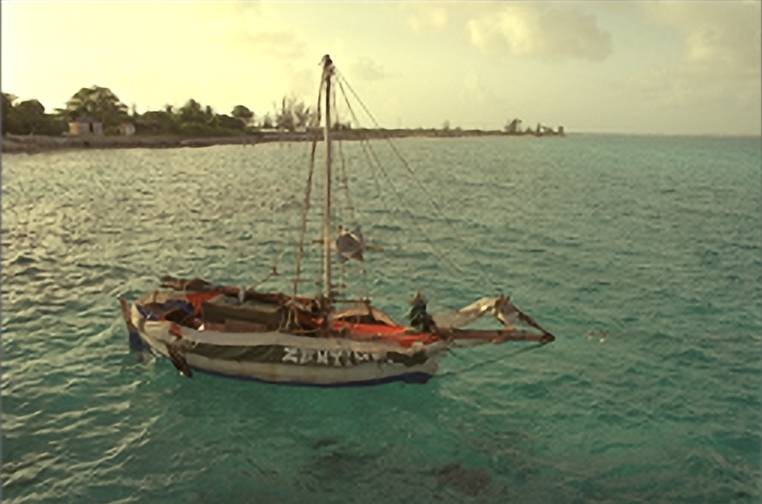} &
      \includegraphics[width=0.24\linewidth]{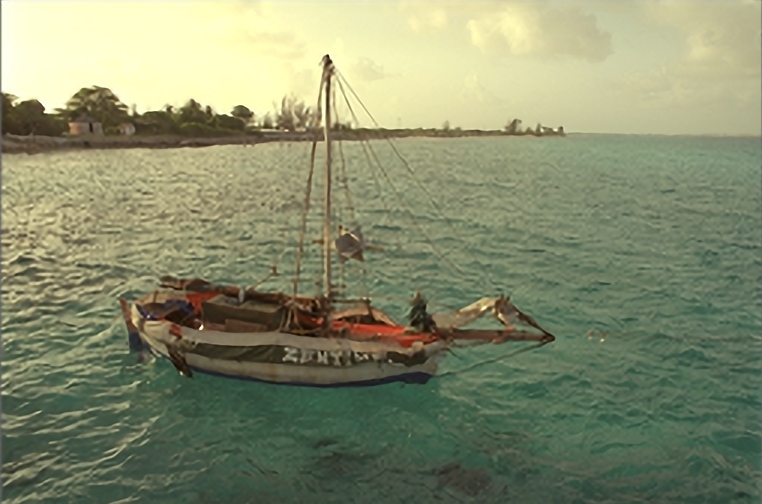} \\
      \includegraphics[width=0.24\linewidth,trim=270 350 360 50,clip]{figures/kodim06.png/fact3_bicubic_kodim06.jpg} &
      \includegraphics[width=0.24\linewidth,trim=270 350 360 50,clip]{figures/kodim06.png/sc.png} &
      \includegraphics[width=0.24\linewidth,trim=270 350 360 50,clip]{figures/kodim06.png/fact3_srcnn_kodim06.png} &
      \includegraphics[width=0.24\linewidth,trim=270 350 360 50,clip]{figures/kodim06.png/fact3_exp_kodim06.png} \\
      Bicubic & Sparse coding~\cite{zeyde2010single} & CNN2~\cite{dong2015image} & SCKN (Ours)
   \end{tabular}
   \caption{Another visual comparison for x3 image up-scaling. See caption of Figure~\ref{fig:visual}.
Best seen by zooming on a computer screen with an appropriate PDF viewer that does not smooth the image content.
}\label{fig:visual2}
\end{figure} 

\begin{figure}
   \renewcommand\tabcolsep{0.1cm}
   \begin{tabular}{cccc}
      \includegraphics[width=0.24\linewidth]{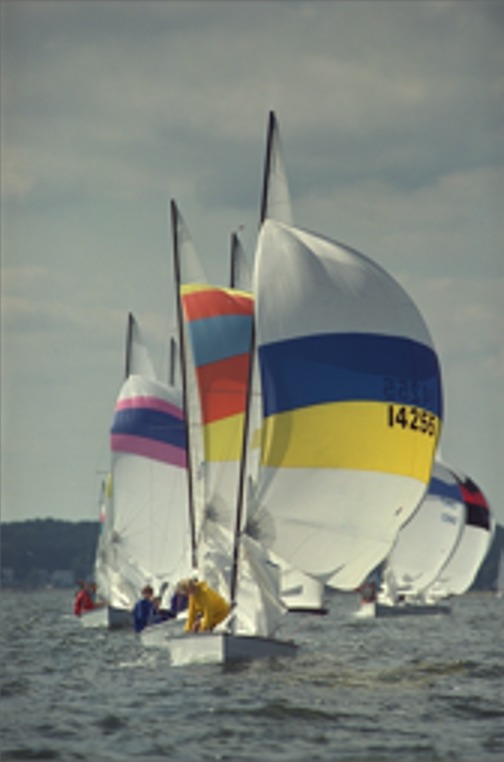} &
      \includegraphics[width=0.24\linewidth]{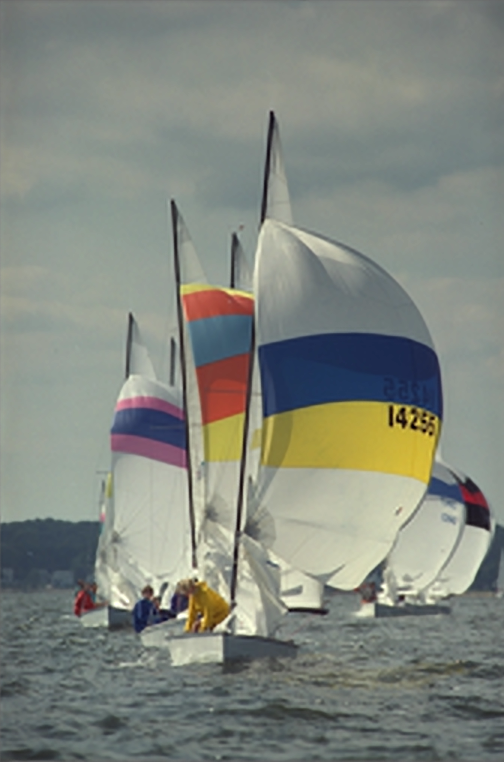} &
      \includegraphics[width=0.24\linewidth]{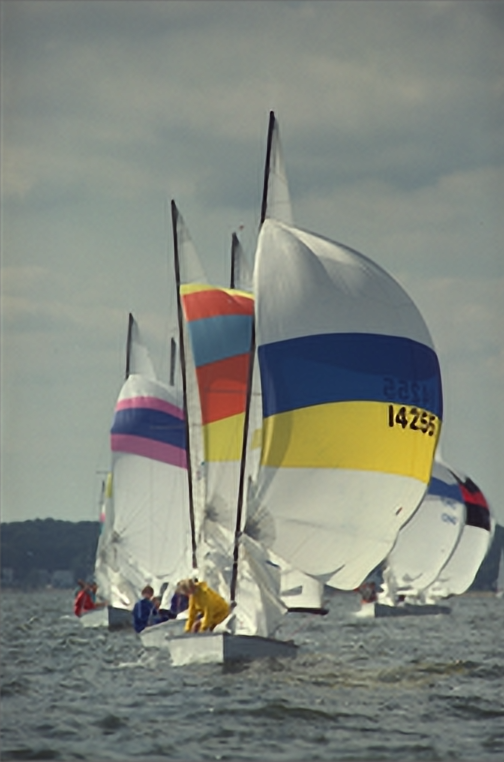} &
      \includegraphics[width=0.24\linewidth]{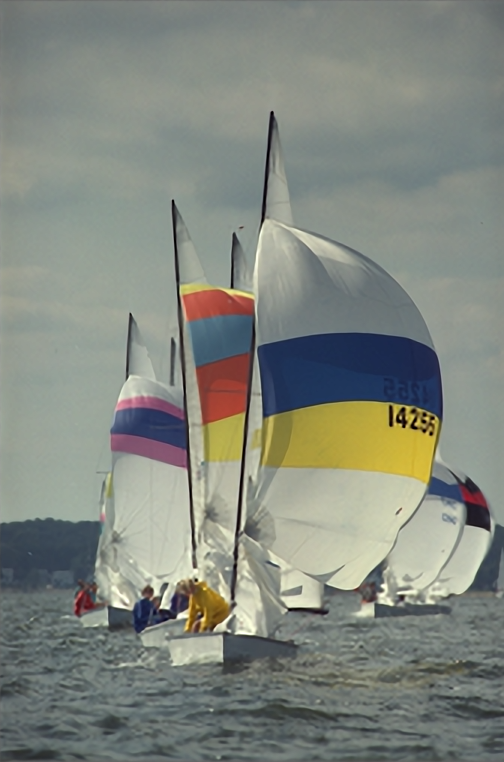} \\
      \includegraphics[width=0.24\linewidth,trim=250 610 210 100,clip]{figures/kodim09.png/fact3_bicubic_kodim09.jpg} &
      \includegraphics[width=0.24\linewidth,trim=250 610 210 100,clip]{figures/kodim09.png/sc.png} &
      \includegraphics[width=0.24\linewidth,trim=250 610 210 100,clip]{figures/kodim09.png/fact3_srcnn_kodim09.png} &
      \includegraphics[width=0.24\linewidth,trim=250 610 210 100,clip]{figures/kodim09.png/fact3_exp_kodim09.png} \\
      \includegraphics[width=0.24\linewidth]{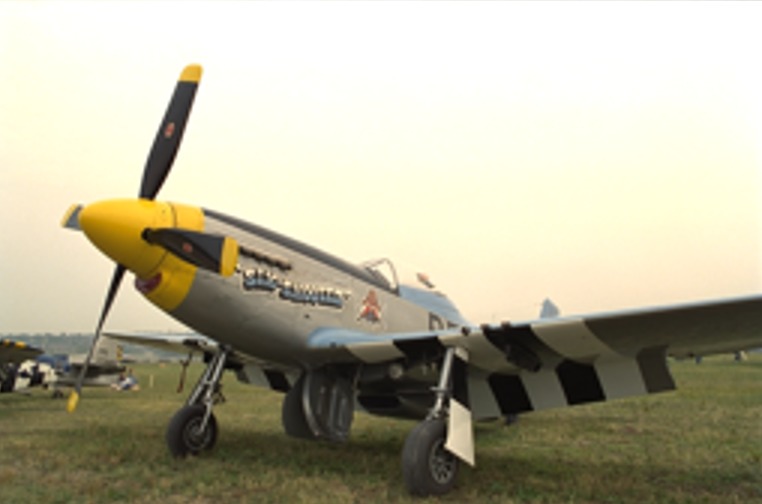} &
      \includegraphics[width=0.24\linewidth]{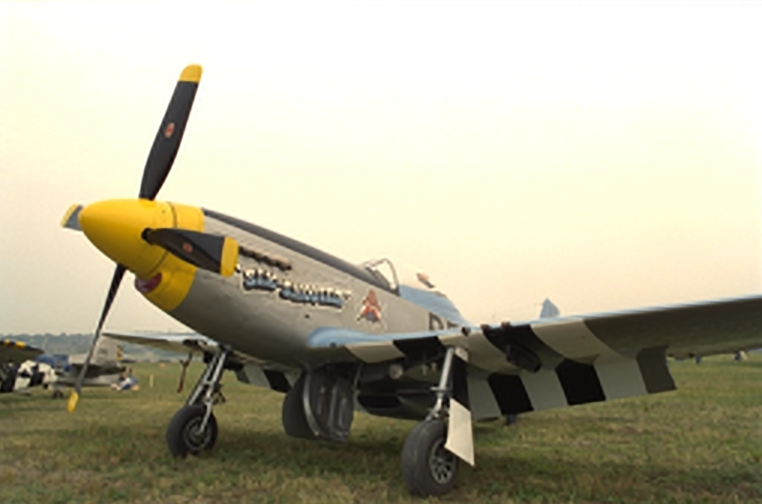} &
      \includegraphics[width=0.24\linewidth]{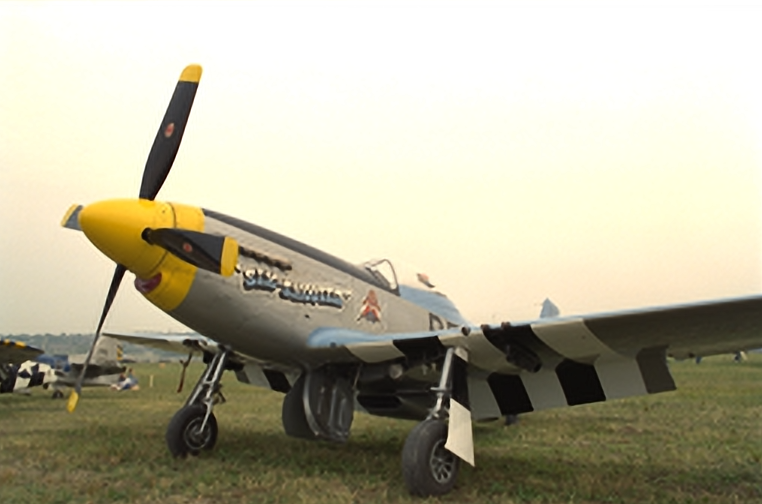} &
      \includegraphics[width=0.24\linewidth]{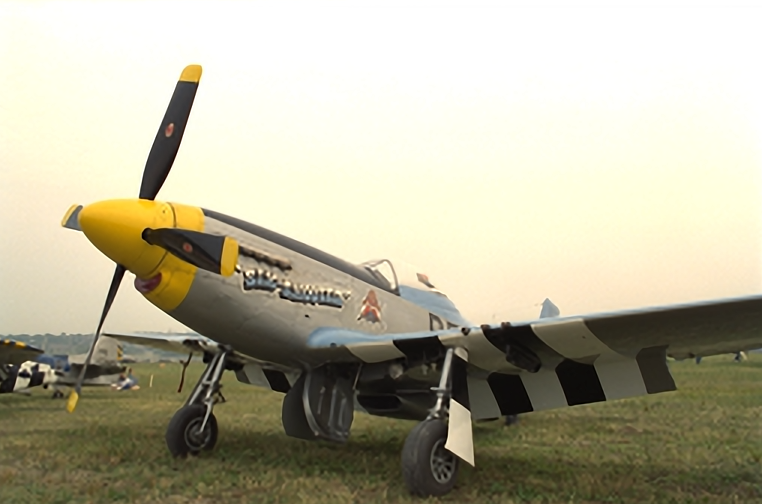} \\
      \includegraphics[width=0.24\linewidth,trim=40 130 510 210,clip]{figures/kodim20.png/fact3_bicubic_kodim20.jpg} &
      \includegraphics[width=0.24\linewidth,trim=40 130 510 210,clip]{figures/kodim20.png/sc.png} &
      \includegraphics[width=0.24\linewidth,trim=40 130 510 210,clip]{figures/kodim20.png/fact3_srcnn_kodim20.png} &
      \includegraphics[width=0.24\linewidth,trim=40 130 510 210,clip]{figures/kodim20.png/fact3_exp_kodim20.png} \\
      \includegraphics[width=0.24\linewidth]{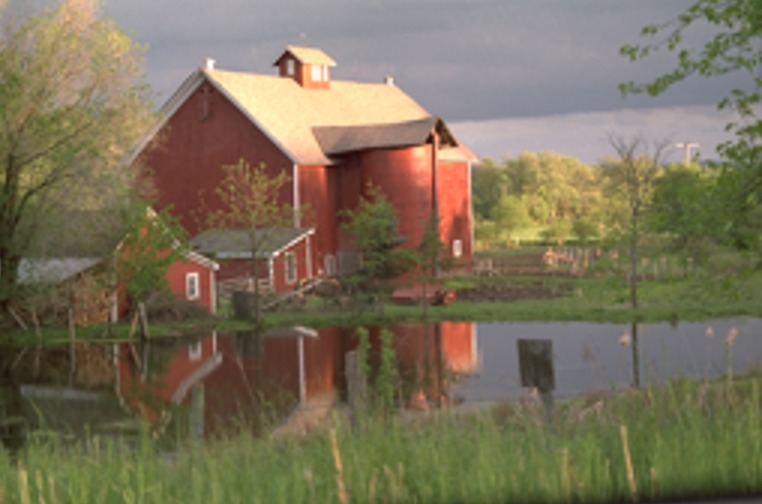} &
      \includegraphics[width=0.24\linewidth]{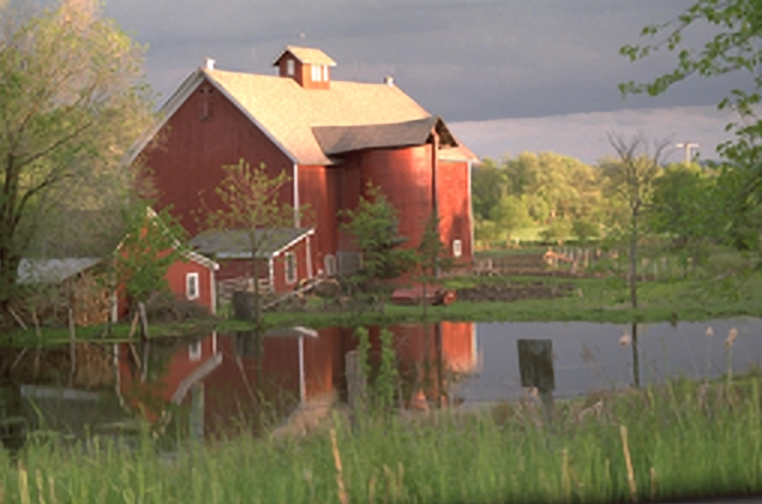} &
      \includegraphics[width=0.24\linewidth]{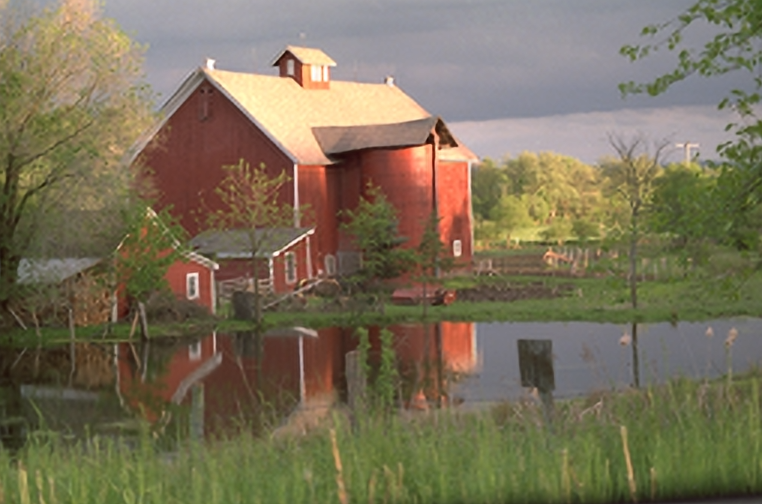} &
      \includegraphics[width=0.24\linewidth]{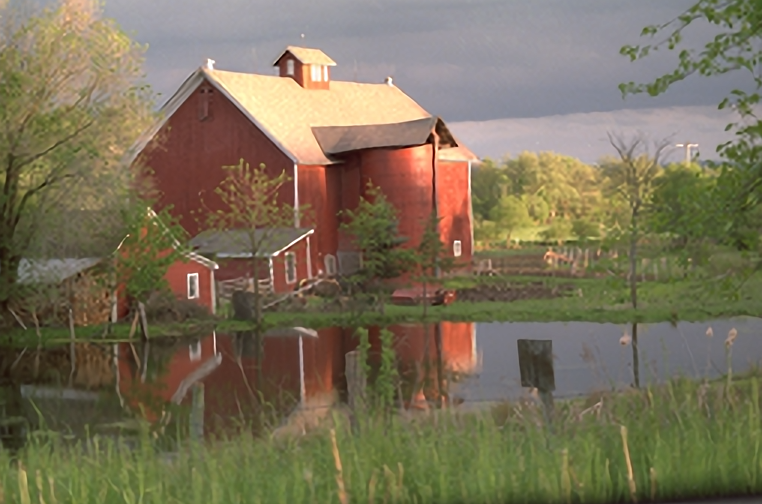} \\
      \includegraphics[width=0.24\linewidth,trim=150 340 400 20,clip]{figures/kodim22.png/fact3_bicubic_kodim22.jpg} &
      \includegraphics[width=0.24\linewidth,trim=150 340 400 20,clip]{figures/kodim22.png/sc.png} &
      \includegraphics[width=0.24\linewidth,trim=150 340 400 20,clip]{figures/kodim22.png/fact3_srcnn_kodim22.png} &
      \includegraphics[width=0.24\linewidth,trim=150 340 400 20,clip]{figures/kodim22.png/fact3_exp_kodim22.png} \\
      Bicubic & Sparse coding~\cite{zeyde2010single} & CNN2~\cite{dong2015image} & SCKN (Ours)
   \end{tabular}
   \caption{Another visual comparison for x3 image up-scaling. See caption of Figure~\ref{fig:visual}.
Best seen by zooming on a computer screen with an appropriate PDF viewer that does not smooth the image content.
}\label{fig:visual3}
\end{figure}

\end{document}